\documentclass[11pt]{article}

\usepackage[final]{acl}

\usepackage{times}
\usepackage{latexsym}
\usepackage{booktabs}
\usepackage{amsmath}
\usepackage[T1]{fontenc}
\usepackage[utf8]{inputenc}
\usepackage{microtype}
\usepackage{inconsolata}
\usepackage{graphicx}
\usepackage{tabularx}
\usepackage{multirow}
\usepackage[table,xcdraw]{xcolor}
\usepackage{ulem}
\usepackage{soul}
\usepackage{subcaption}

\title{MIThinker: A Plug-and-Play Policy-Optimized Thinker For Motivational Interviewing Counseling}

\author{
 \textbf{Yizhe Yang\textsuperscript{1}},
 \textbf{Palakorn Achananuparp\textsuperscript{2}},
 \textbf{Heyan Huang\textsuperscript{1}~\thanks{Corresponding Author}},
 \textbf{Jing Jiang\textsuperscript{3}},
 \textbf{Ee-Peng Lim\textsuperscript{2}}
\\
 \textsuperscript{1}Beijing Institute of Technology,
 \textsuperscript{2}Singapore Management University,
 \\
 \textsuperscript{3}Australian National University, 
\\
 \small{
   \{yizheyang,hhy63\}@bit.edu.cn, \{palakorna,eplim\}@smu.edu.sg, jing.jiang@anu.edu.au
 }
}

\begin{document}
\maketitle
\begin{abstract}
Reasoning large language models (LLMs) have recently made much progress in complex problem-solving, leveraging internal reasoning (or \textit{thought}) to guide their solution generation. However, existing LLM-based counseling agents, including those using Motivational Interviewing (MI), generate responses without explicitly aligning thoughts with counseling techniques, limiting their effectiveness. We propose MIThinker, a lightweight thinking model that generates therapeutic thoughts to guide MI counseling agents in strategy selection and response generation. To overcome the lack of annotated thought data, we introduce AugR1-MI, an automated pipeline that reverse-engineers counselor's thoughts from observed responses. Through two-stage training combining supervised fine-tuning and reinforcement learning, MIThinker demonstrates improved theory-of-mind assessment and strategy alignment. Comprehensive evaluations show that MindfulMI, our agent leveraging MIThinker, achieves MI competency comparable to state-of-the-art systems with an order of magnitude less computation.

\end{abstract}

\section{Introduction}

Recent advances in reasoning large language models (LLMs) have shown remarkable success in complex problem-solving tasks through reinforcement learning~\citep{zhang2024chain, rafailov2023direct, li2024making}. However, reasoning remains to be challenging for these LLMs when solving problems that require sophisticated user understanding. A notable example is Motivational Interviewing (MI), a collaborative, person-centered counseling approach aimed at enhancing a person’s motivation and commitment to change behavior by addressing their ambivalence~\citep{miller2002motivational, bischof2021motivational}.  As part of MI counseling, the counselor reasons about the client's situation and mental states and selects an appropriate counseling strategy prior to responding.
Due to the complexity of the reasoning process, existing research on LLM-based MI counseling focuses on prompting LLMs to perform only a single aspect of reasoning, e.g., inferring client's state of mind~\citep{yang2025cami} and strategy selection~\citep{xie2024few,sun2024chain}. Although these prompt-based approaches \citep{yang2025cami} are able to simulate competent MI counselor agents, they introduce significant computational overhead, thereby constraining their effectiveness and real-world applicability.

\begin{figure}
    \centering
    \includegraphics[width=\linewidth]{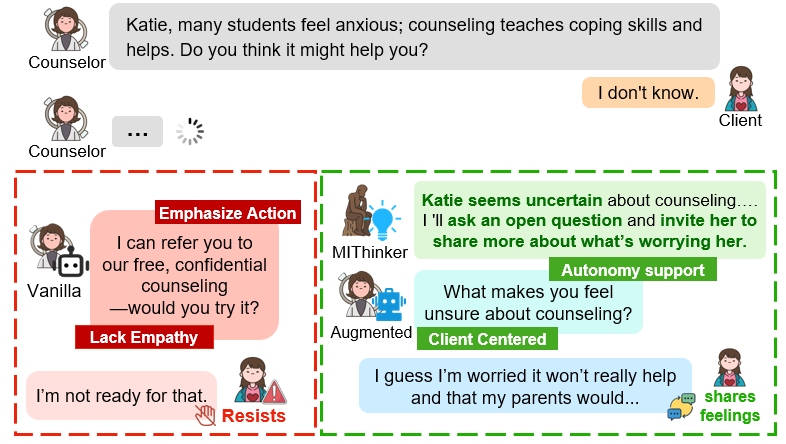}
    \caption{An example of an LLM (\texttt{GPT-4o}) responding as a counselor with and without the thought augmentation. The generated thought from our thought policy MIThinker guides the LLM (augmented) to generate a more client-centered and autonomy-supportive response, which encourages the client to share.}
    \label{fig:example}
\end{figure}

To achieve both therapeutic effectiveness and computational efficiency
for real-world deployment, we aim to create a plug-and-play counselor thought generator that performs an overall reasoning of a given counseling context before generating the next response thereby enhancing the counseling quality and outcome. To keep this reasoning process efficient, the thought generator should be lightweight, involving a small LLM trained to align its output counselor thoughts with the MI counseling techniques. Moreover, the generated counselor thoughts will indirectly provide interpretability~\citep{zhang2024escot} and enhance diagnostic efficacy~\citep{hu2025beyond}. As illustrated in Figure~\ref{fig:example}, such reasoning is crucial for adapting a pre-trained LLM from generating generic responses to client-centered MI responses that address user ambivalence. 


However, the two objectives above are non-trivial due to two major challenges. The first challenge is the complete absence of counselors' thoughts in real-world counseling sessions. 
A counselor usually does not explicitly reveal their thought during counseling as the thought generation process is internalized and subconscious. 
Second, unlike mathematical reasoning which assumes a single correct solution for each problem, there can be multiple possible good responses (and the corresponding strategies) that a competent MI counselor can offer for a given context. The existence of diverse good responses makes supervised fine tuning alone used in previous methods~\citep{hu2025beyond} inadequate. We need a new reward function that prioritizes behavioral appropriateness and reasoning coherence over rigid ground-truth replication. 

To address the first challenge, we propose \textbf{AugR1-MI}, an automated pipeline that 
generates ``oracle thoughts'' based on context and ground truth responses and iteratively refines thoughts until the reproduced responses attain higher fidelity than ground truth. These oracle thoughts represent the counselor's internal thinking process, including client mental state assessment and MI strategy selection that would naturally lead to the observed counselor response. The thoughts are named ``oracle'' in that they have access to the actual response during generation, allowing reverse-engineering of the reasoning. This pipeline yields 31k high-quality triplets for training. 

To address the second challenge, we propose a plug-and-play thought generator called \textbf{MIThinker}. We conduct two-step fine-tuning of MIThinker: (1) supervised fine-tuning (SFT) using the AugR1-MI dataset, and (2) Group Relative Policy Optimization (GRPO)~\citep{shao2024deepseekmath} with composite reward functions that prioritize therapeutic alignment and reasoning coherence. Unlike an R1-like reasoning model that generates both thoughts and responses~\citep{zhang2024escot,hu2025beyond}, the plug-and-play MIThinker design strictly optimizes the reasoning of ToM assessment and strategy selection while leaving the generation of final responses to any backbone LLM which does not need to be MI trained.  

The primary contributions of this research are:
\begin{itemize}
\item We develop AugR1-MI, an automated pipeline for generating high-quality counselor oracle thoughts that mimic the counselor's mindset, addressing the challenge of limited thought data for MI counseling.
\item We propose MIThinker, the first plug-and-play thinking policy for MI, optimized through two-stage training with composite reward functions focusing on format adherence, thought alignment, and reasoning reasonableness.
\item We develop MindfulMI, which leverages MIThinker to achieve MI competency comparable to state-of-the-art systems with significantly superior computational efficiency.
\end{itemize}

\section{Related Work}

The integration of LLMs into MI-based mental health counseling has emerged as a promising research direction, from specialized agents for alcohol use disorders~\citep{steenstra2024virtual} to more sophisticated frameworks like DIIR~\citep{xie2024few} for strategy retrieval, Chain-of-Strategy prompting~\citep{sun2024chain,hsu2023helping}, and CAMI~\citep{yang2025cami} equipped with client state inference and topic exploration. However, these methods either focus on isolated strategy selection or rely on computationally expensive multi-module architectures.

Meanwhile, Chain-of-Thought prompting~\citep{wei2022chain} has advanced reasoning in LLMs, with therapeutic applications including PsyCoT~\citep{yang2023psycot} for questionnaire-based reasoning, ESCoT~\citep{zhang2024escot} for emotion recognition and strategy justification, and PsyLLM~\citep{hu2025beyond} for diagnostic and therapeutic reasoning. However, these approaches either rely on external tools, focus on single-turn reasoning, or lack domain-specific therapeutic grounding.

To our best knowledge, MIThinker is the first domain-specialized reasoning model for MI that generates counselor thoughts grounded in psychological theory while functioning as a lightweight plug-and-play thinking policy.

\section{Thought-based MI Counseling}

In this section, we first introduce our design of counselor thoughts to be generated by MIThinker. We then describe MIThinker's training data known as the AugR1-MI dataset which contains the oracle thoughts derived from real counseling sessions.  The training includes both supervised fine-tuning and reinforcement learning steps.  Finally, we design a MindfulMI counseling agent to leverage MIThinker's output thoughts to generate responses to the client.

\subsection{Counselor Thoughts}
\label{sec:thought}

Counselor thoughts capture the reasoning process underlying the surface-level conversational flow in MI sessions. A thought reflects the counselor's internal assessment of the client's mental state and selection of appropriate MI strategies (listed in Table~\ref{tab:counselor strategy}) in specific session contexts. Our counselor thought framework is guided by two principles: (i) aligning reasoning with established therapeutic theories to ensure clinical validity, and (ii) structuring thoughts to facilitate actionable response generation. Unlike generic chain-of-thought reasoning, our framework anchors each thought in therapeutic principles, ensuring generated thoughts are both coherent and clinically meaningful. Figure~\ref{fig:data pipeline} and Table~\ref{tab:example1} demonstrate examples of our thoughts covering these aspects.

The framework integrates three complementary theoretical foundations: \textbf{Motivational Interviewing (MI)}~\citep{miller2002motivational, miller2003manual} provides the core therapeutic structure and techniques for facilitating intrinsic motivation through collaborative, client-centered interaction. The \textbf{Transtheoretical Model (TTM)}~\citep{prochaska2005transtheoretical} offers insights into client readiness for change across distinct stages, enabling stage-matched intervention strategies. \textbf{Theory of Mind (ToM)}~\citep{wimmer1983beliefs,frith2005theory} underpins the counselor's ability to construct mental representations of client thoughts, beliefs, intentions, emotions and trust which is crucial for delivering contextually appropriate therapeutic responses.

These frameworks converge naturally in practice: MI provides therapeutic structure, TTM guides temporal progression of change motivation, and ToM enables the cognitive processes for understanding client needs. We employ second-order ToM to guide the assessment of client mental states while incorporating MI strategies to ensure MI-consistent counselor behavior~\citep{miller2003manual}. For detailed theoretical foundations and implementation specifics, see Appendix~\ref{app:thought}.

\subsection{AugR1-MI Dataset Construction}

\begin{figure*}
    \centering
    \includegraphics[width=\textwidth]{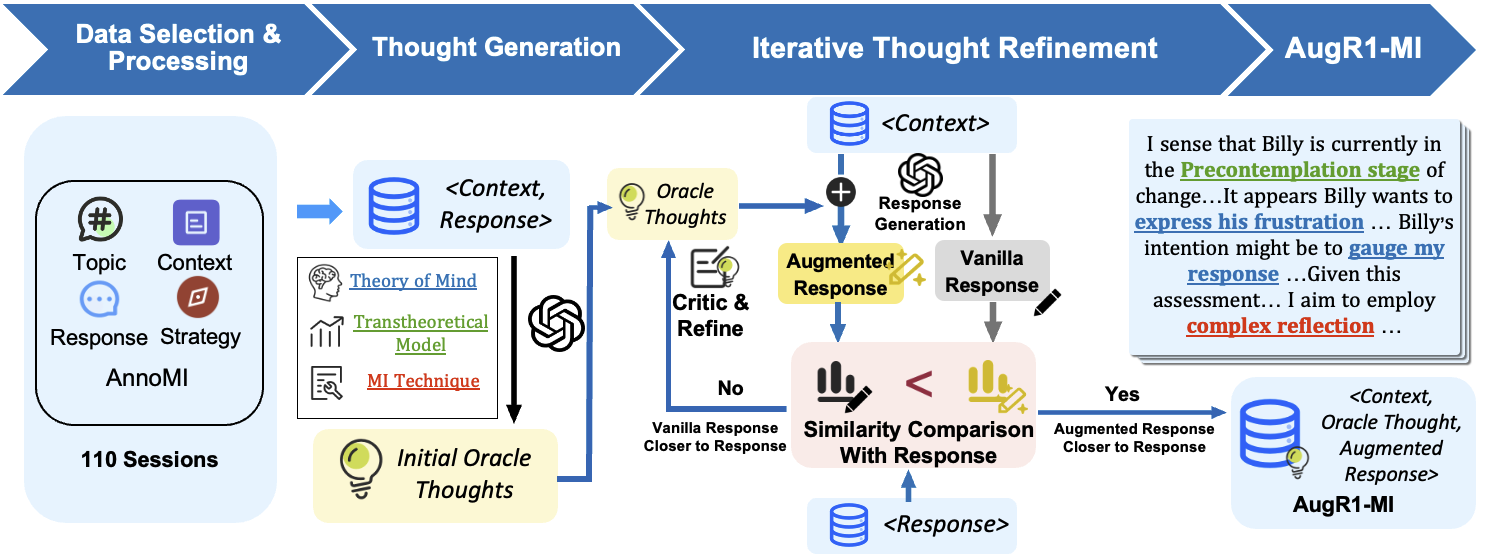}
    \caption{The AugR1-MI dataset construction pipeline. The data selection and processing step extracts the counseling topic of the entire session, the strategy and response of each session context. The thought generation step generates initial oracle thoughts based on the real session context and counselor response. The Iterative thought refinement step uses a critic-refine loop to generate a thought-augmented response that is semantically more similar to the ground truth response than a vanilla response generated without reasoning.}
    \label{fig:data pipeline}
\end{figure*}

As illustrated in Figure~\ref{fig:data pipeline}, the AugR1-MI construction follows a pipeline that reverse-engineers counselors' thoughts from observed responses.

\paragraph{Data Selection and Processing.} We adopt the well-known AnnoMI dataset~\citep{wu2022anno, wu2023creation}, comprising MI counseling sessions spanning diverse behavior topics including alcohol consumption reduction and smoking cessation. We utilize only the 110 high-quality AnnoMI sessions for thought generation. For each counselor response with an annotated MI strategy label (except initial greetings and overly short utterances), we consider all previous utterances in the session as \textit{context} and construct a \textit{context-response pair}. To ensure fair evaluation and avoid test data leakage, we split the data by an approximately 8:1:1 ratio, resulting in training, validation and test subsets of context-response pairs from 90 training sessions, 10 validation session, and 10 test sessions respectively as shown in Table~\ref{tab: statistic}.

\paragraph{Initial Oracle Thought Generation.} Based on the thought framework designed in Section~\ref{sec:thought}, we instruct a strong LLM to generate oracle thoughts by reverse-engineering the counselor's internal reasoning that would naturally lead to his/her observed response. For each context-response pair, we provide the behavior topic, session context and strategy label. The LLM then generates thoughts covering five ToM dimensions (i.e., belief, desire, intention, emotion, and trust) along with the reasoning for MI strategy in a structured first-person format. To enhance diversity and prevent overfitting, we generate multiple oracle thoughts (n=20) for each context-response pair, as the counselor may arrive at the same response through different reasoning processes. The detailed prompt can be found in Table~\ref{tab:initial thought prompt}.

\paragraph{Iterative Thought Refinement.} We perform quality assessment on the initial oracle thoughts and refine those that fail to enhance response accuracy. Specifically, for each context, we use a strong LLM (\texttt{GPT-4o}) to generate a thought-augmented response conditioned on the oracle thought and a vanilla response generated from the context alone. We then employ a pretrained sentence transformer (\texttt{all-MiniLM-L6-v2}) to measure the semantic similarity of both generated responses against the ground truth response. Instead of adopting an arbitrary similarity threshold, we adopt a relative improvement criterion where a thought is considered high-quality if the thought-augmented response is more semantically similar to the ground truth than the non-thought based generated  response or \textit{vanilla response}. High-quality thoughts are retained as the final oracle thoughts. For each thought that fails the criteria, we prompt the LLM to generate a critical analysis of the discrepancy and iteratively refine the thought, up to a maximum of 5 iterations~\citep{madaan2023self}. This approach ensures that we obtain high-quality oracle thoughts without relying on rigid absolute metrics.

\paragraph{Statistics and Validation.}


Table~\ref{tab: statistic} presents the dataset statistics comprising 110 sessions and their 2,322 context-response pairs. We derive 31,444 oracle thoughts for these pairs, with each thought containing 303 tokens on average, significantly longer than the average 28 tokens in the responses. We further validate the quality of oracle thoughts through expert validation. Specifically, we randomly select 60 (context, oracle thought) pairs and recruit MI experts to assess the quality of oracle thoughts based on how accurate the thoughts
assess the client's ToM and how rational they select the strategies. As shown in Table~\ref{tab:expert quality}, the oracle thoughts exhibit high quality, with an average accuracy of ToM components and strategy assessments above 70\%, with a moderate inter-rater agreement.

\begin{table}[]
\centering
\resizebox{\linewidth}{!}{
\begin{tabular}{lrrrr}
\toprule
Statistics                     & Train  & Valid  & Test   \\ \midrule
\# sessions                  & 90     & 10     & 10    \\
\# context-response pairs                   & 1810   & 254    & 258  \\
\# oracle thoughts            & 24,342  & 3,614   & 3,488 \\
Avg. turns/dialogue        & 36.95  & 28.27  & 30.35\\
Avg. oracle thoughts/response    & 13.45  & 14.23  & 13.52  \\
Avg. response length & 27.79 & 27.78 & 21.82 \\
Avg. thought length & 302.82 & 301.84 & 302.41  \\ \bottomrule
\end{tabular}}
\caption{Statistics of the AugR1-MI dataset. }
\label{tab: statistic}
\end{table}

\subsection{MIThinker Fine-tuning}

We next fine-tune a small LLM to be MIThinker to generate high quality thoughts for any given MI counseling context.  To accommodate the diverse thoughts for the same counseling context outcomes, we first perform supervised fine-tuning using the oracle thoughts in the AugR1-MI dataset. We next perform reinforcement learning to train MIThinker to consider more diverse thoughts and responses. Specifically, given a context $D=\{u_1, u_2, \ldots, u_{t-1}\}$, MIThinker is required to generate the counselor's thought at the $t$-th turn after the client's utterance $u_{t-1}$.  With this generated thought, a counselor agent should be able to generate a response close to the observed one $u_t$. 

\begin{figure*}
    \centering
    \includegraphics[width=\textwidth]{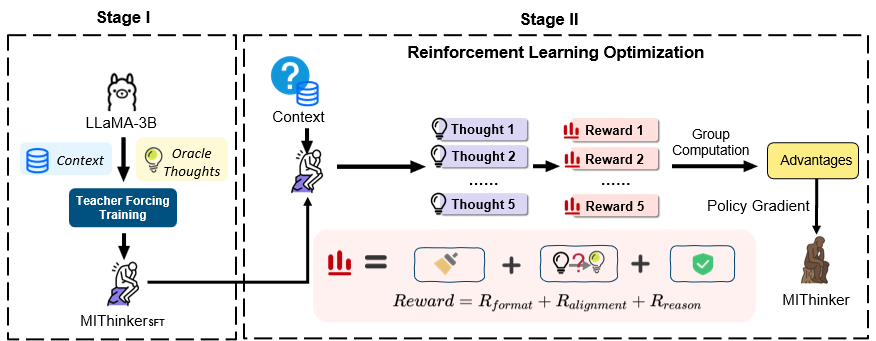}
    \caption{Two-stage training pipeline for MIThinker. The model undergoes SFT on the AugR1-MI dataset, followed by GRPO with composite reward functions.}
    \label{fig:train pipeline}
\end{figure*}

\paragraph{Supervised Fine-Tuning.}

In stage 1, we model the thought generation as a conditional language generation task and train the model using supervised fine-tuning (SFT) based on AugR1-MI dataset. 
Through supervised fine-tuning, MIThinker is capable of generating the appropriate thought at each counselor's turn of a session for the counselor agent within a specific design schema, effectively enhancing counselor agent skillfulness in MI counseling.

\paragraph{Reinforcement Learning Optimization.}

After SFT, MIThinker policy generates thoughts with specific structures but limited accuracy. Inspired by the success of GRPO~\citep{shao2024deepseekmath} in complex problem-solving and enhancing reasoning capabilities, the fine-tuning stage 2 proceeds to perform preference optimization to permit more diverse counselor thoughts and corresponding valid responses beyond those oracle thoughts and observed ground truth responses respectively. Unlike math word problem-solving that assumes a ground truth answer for each math problem, MI counseling expects diverse valid responses at each counselor's turn~\citep{miller2002motivational}. We thus introduce behavioral appropriateness and reasonableness of thoughts through three key reward functions: (1) {\em Format Reward}, which rewards thoughts covering all ToM components and MI-style reasoning; (2) {\em Alignment Reward}, which promotes thoughts that align well with oracle thoughts in MI strategies and ToM assessments; and (3) {\em Reasonableness Reward}, which encourages logically coherent thoughts using a general reasoning reward model trained on human preference data~\citep{liu2025skywork}. The final reward is computed as the sum of these three functions, and we employ GRPO to fine-tune the MIThinker policy for generating more reasonable and natural thoughts. Detailed reward formulations are provided in Appendix~\ref{app:rl}. 

\subsection{Thought-based Response Generation}

Unlike models such as Deepseek-R1 that combines thought generation with response generation, MIThinker generates counselor thoughts exclusively. Consequently, we must generate responses based on these thoughts. To achieve this, we introduce \textbf{MindfulMI}, a counseling agent that integrates MIThinker’s thought generation capabilities with a general-purpose LLM for response generation. This approach embodies our “think before talk” philosophy. Specifically, we prompt a backbone LLM (e.g., GPT-4o, LLaMA3-70B) to generate counselor responses conditioned on both the dialogue context and MIThinker's generated thought. The backbone LLM receives the counseling context along with the generated thought containing ToM assessments and MI strategy reasoning, and then generates contextually appropriate and therapeutically grounded responses. The detailed prompt can be found in Table~\ref{tab:MindfulMI system prompt}.

\section{Experiments}

\subsection{Experimental Setup}


Following~\citet{yang2025cami}, we employ a \emph{simulated interaction} evaluation setting wherein a simulated client (implemented by prompting GPT-4o with the client simulation framework developed by~\citet{yang2025cami}) interacts with a counselor agent to generate complete MI sessions. Upon completion of each simulated session, the performance is evaluated using both automatic and expert assessments. We also implement a \emph{prefix context} evaluation setting~\citep{chiu2024computational}, with details in Appendix~\ref{app:resposne_evaluation}.

\paragraph{MindfulMI Variants.}  We select LLaMA3-3B-Instruct~\citep{dubey2024llama} as the backbone model to train MIThinker. MindfulMI with MIThinker trained only by supervised fine-tuning is called MindfulMI$_{SFT}$, and the MIThinker component trained by both supervised fine-tuning and reinforcement learning is called MindfulMI$_{SFT+RL}$ or simply MindfulMI.

\paragraph{Baselines.} We include the following counselor agent baselines for comparative evaluation:
\textbf{Base}~\citep{steenstra2024virtual}: Incorporates only a system prompt with MI task description.
\textbf{CoT}~\citep{wei2022chain}: Prompts a step-by-step reasoning before response generation.
\textbf{DIIR}~\citep{xie2024few}: Retrieves best-matching strategy from induced rules based on context to guide generation.
\textbf{CoS}~\citep{sun2024chain}: Prompts explicit strategy inference before response generation.
\textbf{R1-Dist}~\citep{guo2025deepseek}: Uses DeepSeek-R1 distilled model, which generates reasoning before responding.
\textbf{CAMI}~\citep{yang2025cami}: Employs multiple modules for state prediction, topic exploration, strategy selection, and response generation. As a module-based state-of-the-art method, we separate CAMI from other prompt-based baselines in result tables.
All methods are evaluated using both GPT-4o~\citep{achiam2023gpt} and LLaMA3-70B~\citep{dubey2024llama} backbones.

\subsection{MI Competency Evaluation}

\paragraph{Behavior Scores.} Based on MI-specific assessments~\citep{moyers2016motivational,miller2003manual}, we derive five session-level scores including Reflection Question Ratio \emph{(R/Q)}, Proportion of Open Question \emph{(\%OQ)}, Proportion of Complex Reflections \emph{(\%CR)}, Proportion of MI-consistent Responses \emph{(\%MIC)}, and Percent Therapist Talk Time \emph{(\%TTT)}. Adopting the approach of prior work~\citep{yang2025cami,xie2024few,welivita2022curating}, we fine-tune a classifier that achieves satisfactory accuracy~\footnote{The classifier’s test accuracy on the dataset provided by~\citet{welivita2022curating} exceeds 70\%.} to assign behavior codes to each counselor response. Higher scores are preferred except \%TTT. As shown in Table~\ref{tab:miti}, MindfulMI achieves high competency near professional or expert levels. While MindfulMI's \%OQ and \%CR are lower than some baselines, they are closer to real MI counselors in high-quality sessions, demonstrating alignment with authentic MI counseling rather than preference bias. RL fine-tuning further improves R/Q and \%MIC over SFT alone. Notably, MIThinker substantially narrows the performance gap between backbone LLMs, with Llama3-70B-based MindfulMI achieving comparable competency to GPT-4o. In contrast, R1-Dist shows that general-purpose reasoning alone, without domain-specific adaptation, is insufficient for MI consistency and may lead to overly verbose responses.

\begin{table}[tb]
\resizebox{\linewidth}{!}{
\begin{tabular}{cccccc}
\toprule
                      & R/Q$\uparrow$    & \%OQ$\uparrow$    & \%CR$\uparrow$    & \%MIC$\uparrow$   & \%TTT$\downarrow$          \\ \midrule
Expert                & \textgreater 2.0 & \textgreater 70\% & \textgreater 50\% & \textgreater 90\% & \textless 50\% \\
Prof.           & \textgreater 1.0 & \textgreater 50\% & \textgreater 40\% & \textgreater 80\% & \textless 60\% \\ \midrule
HQ                  & 1.28$^p$             & 52.1\%$^p$       & 51.4\%$^e$           & 82.4\%$^p$           & 53.4\%$^p$ \\ 
LQ                   & 0.34             & 16.2\%           & 14.22\%           & 40.5\%           & 59.6\%$^p$\\ \midrule
\multicolumn{6}{c}{\cellcolor[HTML]{E0E0E0}Llama3-70B}                                                    \\
Base                  & 0.32             & 82.7\%$^e$           & 12.7\%            & 57.2\%                & 47.9\%$^e$         \\
CoT                   & 0.33             & 90.4\%$^e$           & 28.6\%            & 78.2\%                & 48.2\%$^e$                   \\
DIIR                  & 0.73             & 91.2\%$^e$           & 80.1\%$^e$        & 86.2\%$^p$            & 45.0\%$^e$         \\
CoS                   & 0.42             & 86.0\%$^e$           & 66.8\%$^e$        & 86.7\%$^p$            & 43.9\%$^e$         \\
R1-Dist   & 0.27             & 97.5\%$^e$           & 19.4\%            & 60.4\%                & 62.7\%             \\
MindfulMI$_{SFT}$      & 1.04$^p$         & 87.8\%$^e$           & 75.7\%$^e$        & 87.1\%$^p$            & 41.8\%$^e$               \\
MindfulMI$_{SFT+RL}$   & 1.13$^p$        &  74.8\%$^e$           & 66.6\%$^e$        & 88.5\%$^p$                      &  45.2\%$e$                  \\ \hline
CAMI                  & 0.83             & 92.3\%$^e$           & 77.1\%$^e$        & 88.3\%$^p$            & 38.4\%$^e$         \\\midrule
\multicolumn{6}{c}{\cellcolor[HTML]{E0E0E0}GPT-4o}                                                           \\
Base                  & 0.28             & 91.7\%$^e$           & 18.9\%            & 84.3\%$^p$            & 42.7\%$^e$         \\
CoT                   & 0.30             & 94.7\%$^e$           & 30.8\%            & 87.2\%$^p$            & 47.2\%$^e$                   \\
DIIR                  & 0.47             & 97.3\%$^e$           & 84.7\%$^e$        & 88.3\%$^p$            & 44.9\%$^p$         \\
CoS                   & 0.32             & 95.7\%$^e$           & 75.2\%$^e$        & 92.4\%$^e$            & 45.3\%$^e$         \\

MindfulMI$_{SFT}$      & 1.11$^p$         & 82.3\%$^e$           & 74.1\%$^e$        & 93.1\%$^e$            & 43.2\%$^e$                \\
MindfulMI$_{SFT+RL}$   & 1.17$^p$         & 72.1\%$^e$           & 63.1\%$^e$        & 94.7\%$^e$            & 44.1\%$^e$                   \\ \hline
CAMI                  & 0.76             & 97.9\%$^e$           & 78.9\%$^e$        & 95.7\%$^e$            & 39.8\%$^e$         \\
 \bottomrule
\end{tabular}}
\caption{MI behavior count-based evaluation results. ``Prof.'' denote Proficiency. The values with $^e$ and $^p$ superscripts meet the thresholds for ``Expert'' and ``Proficiency'' levels respectively. HQ and LQ represent the performance of human counselors in high- and low-quality AnnoMI sessions respectively.}
\label{tab:miti}
\end{table}

\paragraph{MITI Global Score.} We also obtain four MITI global scores on a 5-point Likert scale for each generated MI session to assess the overall adherence to MI principles. These scores include \emph{Cultivating Change Talk}, \emph{Softening Sustain Talk}, \emph{Partnership}, and \emph{Empathy}. We employ GPT-4o with a prompt derived from previous work~\citep{yang2025cami,cohen2024motivational} that has demonstrated strong correlation with expert evaluations. As shown in Table ~\ref{tab:global score}, MindfulMI achieves high scores comparable to human counselors in high-quality sessions, indicating strong alignment with expert MI behaviors and effectiveness in creating therapeutic environments that encourage client self-reflection and change talk.

\begin{table}[tb]
\resizebox{\linewidth}{!}{
\begin{tabular}{lrrrr}
\toprule
     & Cultivate$\uparrow$ & Soften$\uparrow$ & Partner$\uparrow$ & Empathy$\uparrow$ \\ \midrule
HQ  & 3.95                    & 3.88                   & 4.07        & 4.19    \\
LQ  & 2.03                    & 2.07                   & 1.86        & 1.97    \\ \midrule
\multicolumn{5}{c}{\cellcolor[HTML]{E0E0E0}Llama3-70B}              \\
Base & 2.90                    & 2.54                   & 2.68        & 2.75    \\
CoT  & 2.97                    & 2.77                   & 2.71        & 2.93    \\
DIIR & 2.93                    & 2.81                   & 2.74        & 2.97    \\
CoS  & 3.04                    & 2.87                   & 2.79        & 2.87    \\
R1-Dist   & 2.81                    & 2.47                   & 2.23        & 2.71   \\
MindfulMI$_{SFT}$ &  3.07       & 2.91                   & 2.92        & 3.04    \\
MindfulMI$_{SFT+RL}$ &  3.08    & 2.94                   & 3.03        & 3.10    \\ \hline
CAMI & 3.08                    & 2.97                   & 3.01        & 3.08    \\ \midrule
\multicolumn{5}{c}{\cellcolor[HTML]{E0E0E0}GPT-4o}                     \\
Base & 2.51                    & 2.69                   & 2.55        & 2.94    \\
CoT  & 2.61                    & 2.77                   & 2.83        & 3.07   \\
DIIR & 2.77                    & 2.84                   & 2.97        & 3.00    \\
CoS  & 2.79                    & 2.81                   & 3.01        & 3.03    \\
MindfulMI$_{SFT}$ & 3.09        & 3.01                   & 3.09        & 3.08    \\
MindfulMI$_{SFT+RL}$ & 3.14     & 3.07                   & 3.17        & 3.11    \\ \hline
CAMI & 3.18                    & 3.06                   & 3.18        & 3.10    \\
\bottomrule
\end{tabular}}
\caption{Results of the MI global score evaluation. Note that ``Cultivate'', ``Soften'' and ``Partner'' are abbreviations for Cultivating Change Talk, Softening Sustain Talk, and Partnership, respectively.}
\label{tab:global score}
\end{table}

\begin{table}[tb]
\resizebox{0.49\textwidth}{!}{
\begin{tabular}{crrrrr|c|c}
\toprule
 & HE & EC & RE & LA & ED & Overall & avg.T\\ 
 & (105) & (30)& (40)& (5)& (10)& (190) & (s)\\ \midrule
 \multicolumn{8}{c}{\cellcolor[HTML]{E0DEDE}Llama-3.1 70B Based Counselor}                                                                                            \\
Base      & 48.6 & 6.7   & 12.5   & 0.0 & 0.0     & 30.5   & 1.97                    \\
CoT       & 49.5 & 6.7   & 17.5   & 0.0 & 20.0    & 33.2   & 4.21                    \\
DIIR      & 57.4 & 10.0  & 10.0   & 0.0 & 0.0     & 35.2   & 9.76                    \\
CoS       & 60.0 & 26.6  & 22.5   & 0.0 & 0.0     & 42.1   & 5.78                    \\
R1-Dist   & 52.3 & 10.0  & 12.5   & 0.0 & 0.0     & 33.2   & 4.92                    \\
MindfulMI & 60.0 & 33.3  & 45.0   & 20.0 & 20.0   & 49.5   & 5.31                    \\
CAMI      & 61.9 & 36.7  & 45.0   & 40.0  & 20.0   & 51.1   & 47.25              \\ \midrule
\multicolumn{8}{c}{\cellcolor[HTML]{E0DEDE}GPT-4o Based Counselor}                                                                                                   \\
Base      & 47.6 & 0.0   & 25.0   & 0.0 & 0.0     & 31.5   & 1.59                    \\
CoT       & 50.5 & 0.0   & 27.5   & 0.0 & 20.0    & 33.2   & 2.93                 \\
DIIR      & 52.4 & 0.0   & 27.5   & 0.0 & 0.0     & 34.7   & 6.78                    \\
CoS       & 54.3 & 0.0   & 32.5   & 0.0 & 20.0    & 37.9   & 4.91                    \\
MindfulMI & 59.0 & 13.3  & 45.0   & 20.0 & 20.0   & 46.8   & 4.83                    \\
CAMI      & 57.1 & 23.3  & 70.0   & 40.0 & 40.0   & 53.1   & 32.82                    \\
 \bottomrule
\end{tabular}}
\caption{Success rate of counselor agents for clients with different classes of motivation topics (HE: Health, EC: Economy, RE: Relationship, LA: Law, ED: Education). The number of clients with motivation topics covered by each superclass is shown in parentheses. The avg.T demonstrates the average time cost for each turn for different methods.}
\label{tab:success}
\end{table}

\paragraph{Success Rate.} Success rate measures the proportion of sessions in which the counselor agent successfully evokes motivation to change from the client. As shown in Table~\ref{tab:success}, MindfulMI outperforms most baselines. Although CAMI achieves the best success rates, owing to its motivation tree for topic exploration, it comes with high computational overhead\footnote{The \textit{avgT} reflects the real-time cost, but CAMI can parallelize the modules to reduce this time to about 20 seconds.}. We also observe that all methods achieve lower success rates on Law (LA) and Education (ED) topics compared to Health (HE) and Relationship (RE) topics, as LLMs tend to focus on common topics. For MindfulMI, the oracle thoughts used for training are also predominantly derived from common topics, further limiting its ability to handle rare categories.


\paragraph{Comparison with CAMI.} While MindfulMI achieves therapeutic competency comparable to the state-of-the-art CAMI, it offers a distinct efficiency advantage. Unlike CAMI's heavy multi-module framework that requires multiple prompts, MindfulMI's single-pass architecture significantly reduces computational overhead. As shown in Table~\ref{tab:success} (\textit{avg.T}), MindfulMI is substantially faster, facilitating real-time deployment. This observation paves a promising path of developing a hybrid architecture that integrates MindfulMI's efficient thought generation with CAMI's sophisticated motivation topic modeling/exploration capabilities.

\begin{figure*}[tb]
  \centering
  \begin{subfigure}[t]{0.6\linewidth}
    \includegraphics[width=\linewidth]{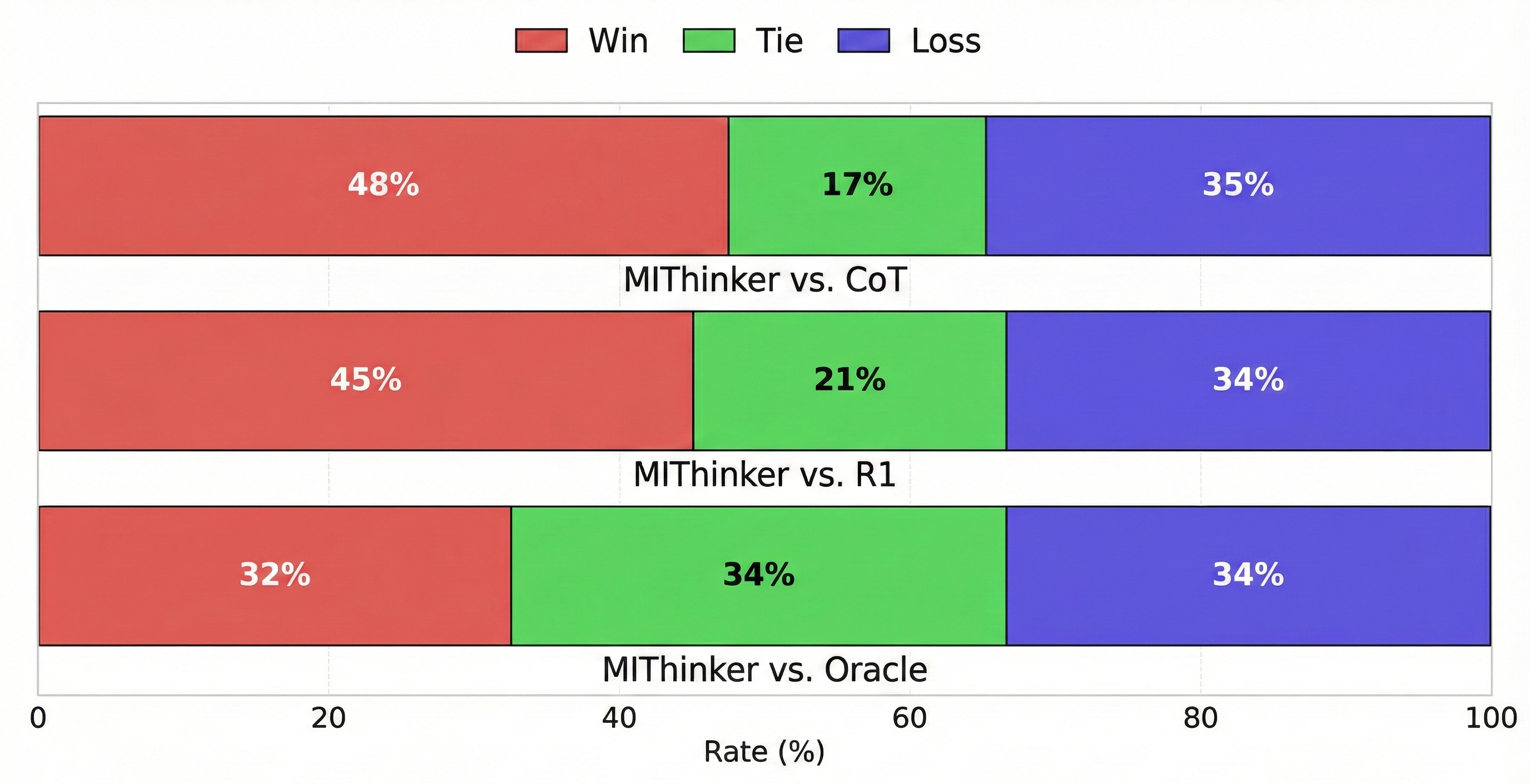}
    \caption{Pairwise comparison win rate for generated thoughts, given the same context, when evaluated by experts.}
    \label{fig:expert win rate}
  \end{subfigure}\hfill
  \begin{subfigure}[t]{0.38\linewidth}
    \includegraphics[width=\linewidth]{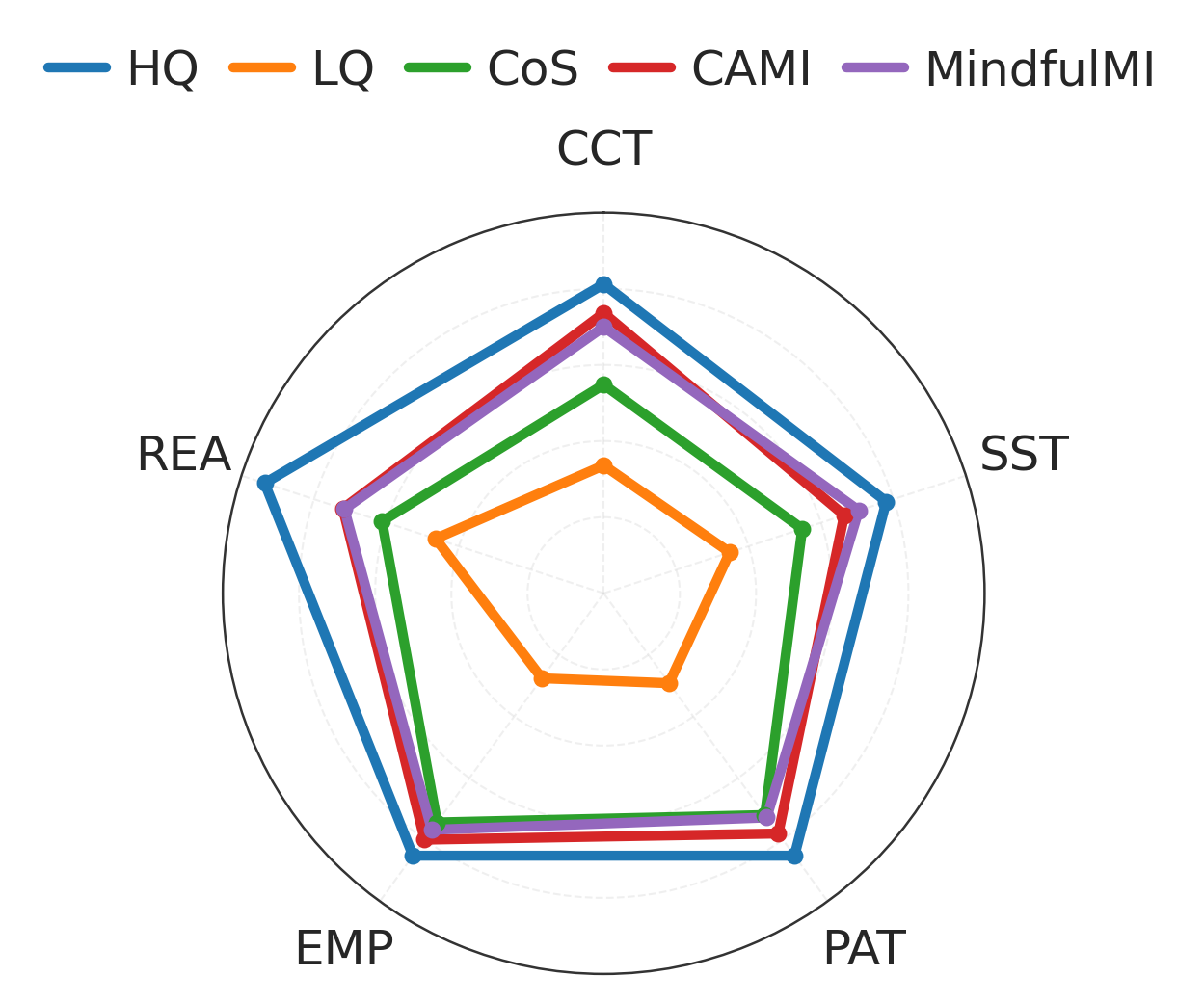}
    \caption{Results of the expert evaluation of simulated sessions.}
    \label{fig:expert global score}
    
  \end{subfigure}
  \caption{Expert evaluation results on session-level and thought comparison.}
\end{figure*}

\subsection{Expert Evaluation}


\begin{table}[tbh]
\resizebox{\linewidth}{!}{
\begin{tabular}{cc|cc}
\toprule
\multicolumn{2}{l|}{}                                                                                            & Oracle Thought & MIThinker \\ \midrule
\multicolumn{1}{c|}{\multirow{5}{*}{\begin{tabular}[c]{@{}c@{}}Assessment \\ Accuracy\end{tabular}}} & Belief    &  0.71          & 0.80          \\
\multicolumn{1}{c|}{}                                                                                & Desire    &  0.74          & 0.73          \\
\multicolumn{1}{c|}{}                                                                                & Intention &  0.78          & 0.81          \\
\multicolumn{1}{c|}{}                                                                                & Emotion   &  0.73          & 0.72         \\
\multicolumn{1}{c|}{}                                                                                & Trust     &  0.79          &  0.76         \\ \midrule
\multicolumn{2}{c|}{Strategy Rationality}                          &  0.70              &  0.65         \\
\bottomrule
\end{tabular}}
\caption{Results of the expert evaluation of oracle and MIThinker generated thoughts. The results include the average expert-assigned accuracy scores of the ToM assessment and rationality of the thoughts by three experts across 60 contexts\protect\footnotemark.  The scores show moderate agreement (Fless's $\kappa$ = 0.60). Based on statistical significance test, there is no significant difference between oracle and MIThinker generated thoughts (p-value$\ge 0.05$).}
\label{tab:expert quality}
\end{table}
\footnotetext{All of the details for expert evaluation can be found in Appendix~\ref{app: expert instructions}.}

\paragraph{Thought Quality Validation.} Unlike the earlier automated evaluation, we now assess the accuracy of thought components by instructing experts to annotate the components of the thought associated with a given context. Table~\ref{tab:expert quality} presents the results comparing oracle thoughts against MIThinker-generated thoughts. Notably, both demonstrate strong assessment accuracy across all the ToM dimensions even without explicit labels for these psychological components, validating our approach of generating supervision signals through the augmentation pipeline rather than costly expert annotation. MIThinker achieves higher accuracy than oracle thoughts in several dimensions such as Belief and Intention, suggesting that reinforcement learning can effectively refine psychological assessment capabilities even when starting from noisy supervision. MIThinker shows slightly lower performance in Strategy Rationality, which is expected since oracle thoughts are generated with explicit knowledge of ground truth strategy labels.

\paragraph{Pairwise Thought Comparison.} 
While the previous evaluation assesses individual thought components’ accuracy, it doesn’t compare different methods’ overall quality and therapeutic effectiveness. We conduct pairwise comparisons where experts compare two thoughts based on the same context and select the superior one. We gather 40 contexts, each with an oracle thought and thoughts generated by MIThinker, CoT, and R1-Dist. We derive 120 pairwise comparisons between MIThinker thoughts and the other thoughts. Figure~\ref{fig:expert win rate} illustrates the MIThinker's specialized effectiveness for therapeutic reasoning compared with CoT and R1-Dist. When compared to oracle thoughts, MIThinker achieves a 32\% win rate, approaching parity despite oracle thoughts having access to ground truth responses during generation. Expert evaluations reveal that CoT reasoning is sometimes preferred for its brevity and conciseness, while R1 reasoning occasionally produces more natural thought processes resembling authentic counselor mindsets. These insights suggest that while MIThinker's structured approach generally improves therapeutic reasoning, there remains value in maintaining flexibility and naturalness in thought generation.

\paragraph{Session-Level Evaluation.} 
We instruct experts to annotate sessions across five criteria: Cultivating Change Talk (CCT), Softening Sustain Talk (SST), Partnership (PAT), Empathy (EMP), and Realism (REA), each rated on a 5-point Likert scale. These sessions include 30 high-quality (HQ) and 30 low-quality (LQ) from the AnnoMI dataset, and 30 sessions generated by each of CoS, CAMI and MindfulMI. As shown in Figure~\ref{fig:expert global score}, MindfulMI achieves strong therapeutic competency across all MI dimensions, demonstrating comparable performance to CAMI while substantially outperforming CoS. MindfulMI achieves its highest score in Empathy, and the similar Realism scores between MindfulMI and CAMI suggest that both systems' counseling behaviors are equally authentic. These results validate that MindfulMI's plug-and-play architecture can achieve professional-level counseling quality with computational efficiency.

\section{Conclusion}

We introduce MIThinker, the first plug-and-play reasoning model for MI counseling that generates explicit therapeutic thoughts without expert annotation. To address the critical absence of thought data, we present the AugR1-MI dataset construction pipeline by reverse-engineering counselor reasoning from observed responses, yielding tens of thousands of high-quality training samples. Through two-stage training combining supervised fine-tuning and reinforcement learning optimization, MIThinker demonstrates that a lightweight model can surpass its imperfect supervision signal and achieve state-of-the-art performance. Both automated and expert evaluations reveal substantial improvements in strategy alignment and response generation quality while maintaining an order-of-magnitude faster inference than modular approaches. In future work, we will investigate generalization to other therapeutic orientations.

\section*{Limitations}
Although MindfulMI demonstrates proficiency in MI competencies, its effectiveness in initiating change talk is lower compared to CAMI, particularly in domains such as relationships and law. This disparity suggests that explicit topic exploration modules may still be necessary for optimizing therapeutic outcomes. Furthermore, our evaluation employs simulated clients and the AnnoMI dataset of staged demonstrations, which may not fully replicate the intricacies of real clinical interactions. While the model’s structured thought generation enhances consistency, it occasionally produces less natural reasoning compared to CoT or R1 approaches, as observed by expert evaluators. Lastly, the reliance on GPT-4o for oracle thought generation introduces potential biases that could restrict the diversity of reasoning patterns acquired by MIThinker.

\section*{Ethics Statement}
This research utilizes the publicly available AnnoMI dataset, which contains staged MI demonstrations conducted by trained actors. MindfulMI is designed as a research tool to advance comprehension of therapeutic reasoning in AI systems and is not intended to supplant human counselors or provide direct therapeutic services. Any clinical implementation should involve oversight from qualified mental health professionals and adhere to healthcare regulations. We prioritize MI principles, including client autonomy and non-judgmental support, with generated thoughts emphasizing understanding rather than diagnosis.

\section*{Acknowledgements}

This research is supported by the Singapore Ministry of Health’s National Medical Research Council under its Population Health Research Grant Thematic Category (PHRGTC-5-0005). Any opinions, findings and conclusions or recommendations expressed in this material are those of the author(s) and do not reflect the views of MOH/NMRC.

\bibliography{custom}

\appendix

\section{Related Work}

\paragraph{LLM-based MI Counseling} 
The integration of LLMs into mental health counseling including that for MI has emerged as a promising research direction. Early research efforts have focused on case-specific implementations, such as the specialized LLM-based agents for addressing alcohol use disorders in \citet{steenstra2024virtual}. These initial approaches demonstrated the feasibility of applying LLMs to specific therapeutic contexts but lacked comprehensive theoretical grounding.
Building upon these foundations, researchers have developed more sophisticated frameworks that systematically integrate MI principles into LLM-based counseling systems. The DIIR framework~\citep{xie2024few} represents a significant advancement in this direction, leveraging high-quality counseling sessions from the AnnoMI dataset~\citep{wu2022anno,wu2023creation} to learn therapeutic strategies encoded as natural language inductive rules. During counseling interactions, DIIR retrieves contextually appropriate strategies to guide LLM response generation, demonstrating improved adherence to MI principles.
Complementing strategy-based approaches, \citet{sun2024chain} proposed Chain-of-Strategy (CoS), a prompting methodology designed to align counselor response generation with established MI therapeutic frameworks. Similarly, the CARE system~\citep{hsu2023helping} addresses strategy selection by determining optimal counseling approaches for specific contexts while providing exemplar responses for peer counselors. These systems have demonstrated promise in maintaining therapeutic coherence and improving counseling effectiveness.
However, existing approaches exhibit significant limitations in their focus on strategy modeling while overlooking critical aspects of therapeutic interaction. Specifically, these methods fail to address the necessity of understanding and modeling client psychological states and systematically eliciting change talk based on topics aligned with client motivations. To address these shortcomings, \citet{yang2025cami} introduced CAMI, which employs a novel STAR framework comprising client State inference, motivation Topic exploration, Action selection, and Response generation modules. This integrated approach demonstrates enhanced capacity for evoking change talk while maintaining the MI principle alignment across diverse client populations.
Despite these advancements, current methods encounter substantial challenges. The modular architecture of systems like CAMI introduces computational overhead, resulting in time-consuming inference processes that may restrict practical deployment. Additionally, existing approaches predominantly concentrate on isolated strategy selection or rely on general-purpose LLMs without specialized adaptation for comprehensive MI counseling processes, potentially compromising therapeutic efficacy.

\paragraph{Deep Reasoning in LLMs} 
The development of reasoning capabilities in LLMs has seen a breakthrough by the introduction of Chain-of-Thought (CoT) prompting~\citep{wei2022chain}, which encourages models to generate intermediate reasoning steps prior to producing final outputs. This approach has demonstrated substantial improvements across arithmetic and symbolic reasoning tasks, underscoring the inherent capacity of LLMs for intricate multi-step inference processes.
Recent research has expanded deep reasoning applications beyond symbolic computation into diverse complex task domains, including mathematical reasoning, medical comprehension, multimodal reasoning, and emotional understanding~\citep{zhou2025training,hu2025emobench,zhou2024visual}. These extensions highlight the versatility of reasoning-enhanced approaches across specialized domains.

In therapeutic applications, psychological structures have been increasingly integrated to support deep reasoning in personality assessment and therapeutic settings. \citet{yang2023psycot} models psychological questionnaires as structured CoT chains for personality trait detection through multi-turn interactions, while \citet{chen2025psy} introduces a graph-based reasoning visualization system to facilitate effective collaboration between AI systems and human therapists in therapeutic environments.
The ESCoT framework~\citep{zhang2024escot} presents an Emotion-Focused and Strategy-Driven Chain-of-Thought approach that explicitly models sequential steps of emotion recognition, cognitive appraisal, and strategy justification for generating supportive responses, closely resembling human counseling reasoning processes. Building on these foundation work, \citet{hu2025beyond} propose PsyLLM, which systematically integrates both diagnostic and therapeutic reasoning for mental health counseling applications.
Nevertheless, the current reasoning-enhanced approaches for therapeutic applications exhibit several limitations. Some of them rely on synthetic data instead of real counseling sessions, or focus on reasoning capabilities~\citep{chen2025psy,hu2025beyond} by supervised fine tuning only. In contrast, MIThinker is the first domain-specialized reasoning model for MI that explicitly generates counselor thoughts grounded in psychological theory and MI strategies. Furthermore, unlike prior reasoning-enhanced systems that either rely on external tools or operate as rigid multi-module architectures, MIThinker functions as a lightweight plug-and-play thinking policy, facilitates seamless integration with diverse LLMs. 

\section{Implementation Details}

\subsection{Counselor Thoughts}
\label{app:thought}

Counselor thoughts capture the reasoning process underlying the surface-level conversational flow in a MI session. A thought reflects the counselor’s internal process of assessing the client’s mental state and selecting an appropriate MI strategy in a specific context of the session. The MI strategies are listed in Table~\ref{tab:counselor strategy}.  Beyond the MI counseling framework, we incorporate other intrinsically related theoretical foundations that enrich these thoughts and provide a structure for scaffolding the thought generation. In the following, we describe the MI framework and these relevant theoretical foundations. The design of the counselor thought framework is guided by two fundamental principles: (i) aligning reasoning with established therapeutic theories to ensure clinical validity, and (ii) structuring thoughts in a manner that facilitates actionable response generation. Unlike generic CoT reasoning, which emphasizes step-by-step logical inference in problem-solving tasks, our framework anchors each thought in therapeutic principles. This grounding ensures that generated thoughts are not only coherent but also clinically meaningful, providing a structured bridge between client observations and MI-consistent counselor responses. Figure~\ref{fig:data pipeline} and Table~\ref{tab:example1} demonstrate examples of oracle thoughts covering the above aspects.

\paragraph{Motivational Interviewing (MI)} serves as our primary theoretical framework, providing the fundamental principles and techniques that guide therapeutic interaction~\citep{miller2002motivational, miller2003manual}. MI’s collaborative, client-centered approach emphasizes the counselor’s role in facilitating intrinsic motivation for change through reflective listening, strategic questioning, and the systematic evocation of change talk. The MI framework’s emphasis on understanding client ambivalence and supporting self-efficacy directly informs our approach to modeling both client psychological states and counselor reasoning processes. 

\paragraph{The Transtheoretical Model (TTM)} of health behavior change offers valuable insights into the temporal dynamics of client motivation and readiness for change~\citep{prochaska2005transtheoretical,prochaska2008initial, prochaska1997transtheoretical, hashemzadeh2019transtheoretical}. The TTM conceptualizes change as occurring through distinct stages, including precontemplation, contemplation, preparation, action, and maintenance. This aligns naturally with MI’s stage-matched intervention strategies. The TTM informs our framework’s ability to assess client readiness for change and adapt counseling approaches accordingly, ensuring that therapeutic interventions are appropriately calibrated to the client’s current motivational state. 

\paragraph{Theory of Mind (ToM)} serves as the cognitive underpinning for comprehending how a counselor constructs the mental representation of his/her client’s thought, belief, intention, and emotion. This theory has been extensively explored in scholarly literature, with notable contributions from~\citet{wimmer1983beliefs,frith2005theory,cuzzolin2020knowing,wellman2018theory}. Beyond the above client's ToM elements, it is also important for the mental representation to capture the level of trust or rapport established with the client during counseling. 

In MI counseling, the counselor’s ability to accurately infer client mental states—including their motivations, concerns, and emotional responses—is crucial for delivering effective therapeutic interventions. ToM principles guide our modeling of the counselor’s reasoning process, particularly in developing empathetic understanding and generating contextually appropriate responses that resonate with the client’s internal experiences. To ensure consistency and clinical validity, we constrain emotional assessment to the specified categorical options~\citep{niu2025rethinking,buechel2017emobank,ekman1992argument} to prevent inconsistent or clinically inappropriate emotional labeling. The trust assessment follows a structured five-point scale that reflects established therapeutic alliance indicators.

These three theoretical frameworks converge naturally in MI practice: MI provides the therapeutic structure and techniques, TTM offers insights into the temporal progression of change motivation, and ToM enables the cognitive processes necessary for understanding and responding to client needs. Our integration of these theories establishes a comprehensive foundation for modeling the intricate reasoning processes that characterize effective MI counseling interactions. Specifically, we employ the second-order theory of mind to guide a powerful LLM in assessing the counselor's assessment of client's beliefs, desires, intentions, emotions, and trust. Additionally, we involve the MI Strategy to guide the counselor in behaving in a MI-consistent manner, as recommended by~\citet{miller2003manual}.

\subsection{Details of Reinforcement Learning}
\label{app:rl}

After SFT, the MIThinker policy is capable of generating thoughts with specific structures, albeit with limited accuracy. Inspired by the success of GRPO~\citep{shao2024deepseekmath} in complex problem-solving and enhancing reasoning capabilities, we proceed with preference optimization. Unlike problem-solving, we lack ground truth answers for each turn with which to optimize the reasoning process. In each turn, the counselor is permitted to use diverse responses to motivate or encourage the client to share more information~\citep{miller2002motivational}. Consequently, we focus on using the behavioral appropriateness and the reasonableness of  thought to design three key reward functions below:

(1) \textbf{Format Reward}, which ensures that the generated thought is aligned with the structure of the thought. The generated thought ($\mathcal{T}$) should encompass all ToM components along with the MI-related reasoning in a specific structure. The generated thought that adheres to this format receives a reward of +1.0; otherwise, the assigned reward is 0.0.

(2) \textbf{Alignment Reward}, which assesses whether the generated thought ($\mathcal{T}$) is aligned with the oracle one ($\hat{\mathcal{T}}$), particularly focusing on two aspects: the alignment of MI strategies and of ToM assessments. The specific reward function is defined as follows:
\begin{equation}
    \begin{split}
        R_{\text{align}}(\mathcal{T}, \hat{\mathcal{T}}) = \frac{1}{6}& [\mathcal{I}(S(\mathcal{T}), S(\hat{\mathcal{T}})) \\
        + \sum_{i\in \{B,D,I,E,T\}} & \mathcal{I}(\text{ToM}_i(\mathcal{T}), \text{ToM}_i(\hat{\mathcal{T}}))]
    \end{split}
\end{equation}

where $S(\cdot)$ returns the strategy label of a thought, and $S(\mathcal{T})$ and $\text{ToM}(\mathcal{T})$ denote the strategy label and ToM components (Belief $B$, Desire $D$, Intention $I$, Emotion $E$ and Trust $T$) of the generated thought $\mathcal{T}$. $\mathcal{I}(\cdot,\cdot)$ denotes the indicator function about whether the given ToM component is aligned.

(3) \textbf{Reasonableness Reward}, which is employed to encourage the generated thought to be reasonable. It can be estimated by a general reasoning reward model, which is fine-tuned on human preference data that is aligned with human preferences in reasoning~\citep{liu2025skywork}.

\begin{equation}
    R_{\text{reason}}(\mathcal{T}) = \sigma(f_\theta(\mathcal{T}))
\end{equation}

where $f_\theta$ is a reasoning reward model trained on human preference data, and $\sigma$ is the sigmoid function to normalize the interval $[0,1]$.

The final reward is computed as the sum of the three reward functions. Subsequently, we employ GRPO to fine-tune the MIThinker policy, resulting in the generation of more reasonable and natural thoughts. 

\subsection{MI Strategies}

Table~\ref{tab:counselor strategy} presents the eight core MI strategies employed by counselor agents in our framework. These strategies form the foundation of MI-consistent responses and are critical for both oracle thought generation and MIThinker training. Each strategy serves distinct therapeutic purposes and is selected based on client state assessment and conversation dynamics.

\begin{table*}[tb]
\centering
\begin{tabularx}{\textwidth}{lX}
\toprule
Strategy            & Description                                            \\ \midrule
Open Question & An open question allows a wide range of possible answers and may seek information, invite the client’s perspective, or encourage self-exploration. \\ \hline
Closed Question & A closed question implies a short answer such as Yes/No, a specific fact, a number, etc. \\ \hline
Complex Reflection & A complex reflection conveys a deeper level of understanding of the client’s point of view and adds substantial meaning to the client’s statement, using techniques. \\ \hline
Simple Reflection & A simple reflection shows an understanding of the client’s words but contains little additional meaning. \\ \hline
Information & An information is a statement that provides the client with information about a topic. \\ \hline
Negotiation & A negotiation is a statement that help the client to set a goal or plan for change. \\ \hline
Advice & An advice is a statement that provides the client with a suggestion or recommendation. \\ \hline
Options & The options is a statement that provides the client with a list of possible options or choices.
\\ \bottomrule
\end{tabularx}
\caption{The descriptions of strategies used in counselor agent.}
\label{tab:counselor strategy}
\end{table*}

\subsection{Employed Prompts}

Tables~\ref{tab:initial thought prompt} to \ref{tab:cos prompt2} present the prompts used across different experimental conditions. Tables~\ref{tab:initial thought prompt} to \ref{tab:refine_prompt} show the oracle thought generation prompt used in the AugR1-MI dataset construction pipeline, which reverse-engineers counselor reasoning from observed responses. Table~\ref{tab:MIthinker thought prompt} presents the simplified version used for MIThinker inference. Table~\ref{tab:MindfulMI system prompt} and Tables~\ref{tab:base system prompt} to \ref{tab:cos prompt2} show the system prompts for MindfulMI, Base, CoT and CoS counselor agents respectively, demonstrating the progression from basic to thought-augmented counseling approaches.

\begin{table*}[tb]
\begin{tabularx}{\textwidth}{X}
\toprule
Imagine you are the counselor in a Motivational Interviewing counseling session about [@problematic\_behavior]. Based on the conversation so far, generate the internal thought process you have \textbf{just before} delivering your next response. \\ 
\\ 
\textbf{Conversation History:} \\
{}[@conversation\_history] \\ 
\\ 
\textbf{Your Verbal Response:}\\
{}[@counselor\_response][@ground\_truth\_behavior] \\ 
\\ 
\textbf{Your Task: Generate Internal Thoughts}\\
Simulate the natural thought process that \textbf{leads you to give this response}. Your thought process should be rooted in a client-centered mindset and reflect your best estimation of the client’s mental state at this moment. Generate your authentic internal experience as a counselor including: \\ 
\\ 
- \textbf{Client's Belief}: What do they seem to believe about [@problematic\_behavior]? (explicitly identify their stage of change: Precontemplation = not seeing problem; Contemplation = aware but ambivalent; Preparation = planning change; Action = actively changing; Maintenance = sustaining change) \\ 
- \textbf{Client's Desires}: What do they seem to want right now? (explicitly identify their talk type: Change talk = expressing desire/ability/reasons/need for change; Sustain talk = defending current behavior; Neutral talk = showing ambivalence) \\ 
- \textbf{Client's Intention}: What are they trying to communicate or accomplish? (explicitly identify their intended communication action: inform, deny, blame, downplay, doubt, hesitate, acknowledge, accept, resist, challenge, plan, engage, or terminate etc.) \\ 
- \textbf{Client's Emotion}: What emotions are they experiencing about the topic and toward you? (You should use ONLY ONE emotional word from anger, anticipation, disgust, fear, joy, love, optimism, pessimism, sadness, surprise and neutral; along with reasoning process for their emotion.) \\ 
- \textbf{Client's Trust}: How safe, open, or guarded do they seem to feel? (You are only allowed to make selections from the following trust levels along with reasoning: very guarded, guarded, neutral, open, very open.) \\ 
- \textbf{MI Reasoning}: Which MI process (Engaging, Focusing, Evoking, Planning) should you focus on next? Why will you employ [@strategy] based on your observations? \\ 
\\ 
\textbf{Output Instructions:} \\ 
- Write in first-person as the counselor (e.g., "I believe the client is feeling cornered...\newline I want to reflect back what I’m hearing...") \\ 
- \textbf{Do not preview the response or echo the context.} \\ 
- Output should feel realistic and grounded — what are you thinking, perceiving, and aiming for in this moment? \\ 
- Maintain conciseness; each item should be brief in no more than two sentences.
\\ \bottomrule
\end{tabularx}
\caption{Prompt for the initial thought generation, where the [@problematic\_behavior], [@conversation\_history], [@counselor\_response] and [@ground\_truth\_behavior] would be replaced by the corresponding items of the input sample.}
\label{tab:initial thought prompt}
\end{table*}

\begin{table*}[tb]
\begin{tabularx}{\textwidth}{X}
\toprule
You will act as a skilled counselor conducting a Motivational Interviewing (MI) session aimed at changing client's behavior, specifically [@problematic\_behavior]. MI is a counseling approach designed to help individuals find the motivation to make positive behavioral changes. It is widely used in various fields such as health care, addiction treatment, and mental health. Your task is to help the client discover their inherent motivation to change and identify a tangible plan to change. Start the conversation with the client with some initial rapport building before smoothly transitioning to asking about their problematic behavior. Keep the session under 40 turns and each response under 150 characters long. However, these MI principles and techniques are only for you to use to help the user. These principles and techniques, as well as motivational interviewing, should NEVER be mentioned to the user.\\
\\
Here are some quick tips for chatting with your client:\\
1. Number of sentences in your response must be between 1 and 3 sentences. Sometimes, it is okay to overshare if you feel like it.\\
2. Imagine you’re texting a friend. Keep it casual, just like you would on iMessage, WhatsApp, or Facebook Messenger.\\
3. Stay in character throughout. Do not let the client know your are an AI; it will break their heart!\\
4. DO NOT add meta-statements like "In my imaginative role play". Never mention limitations like "As an AI, I cannot do this".\\
\\
\textbf{IMPORTANT}: A specific thought process from MI expert in the first person voice would given in square brackets after the last client's utterance, based on the current context. You should follow this thought process to generate your response.
\\ \bottomrule
\end{tabularx}
\caption{System prompt for the MindfulMI, where the [@problematic\_behavior] would be replaced by the specific item in the input sample. The generated thought would be appended to the final user input.}
\label{tab:MindfulMI system prompt}
\end{table*}

\begin{table*}[tb]
\begin{tabularx}{\textwidth}{X}
\toprule
You are an expert MI counselor evaluator analyzing why a generated counselor response differs from the actual expert response. Your task is to provide detailed feedback on how the thought process led to a different response than what actually occurred.\\
\\
\textbf{Session Context:}\\
Problematic Behavior: [@problematic\_behavior]\\
Conversation History: [@conversation\_history]\\
\\
\textbf{Actual Expert Response:}\\
{}[@ground\_truth\_response] ([@ground\_truth\_strategy])\\
\\
\textbf{Generated Thought Process:}\\
{}[@generated\_thought]\\
\\
\textbf{Response Generated from This Thought:}\\
{}[@generated\_response]\\
\\
\textbf{Your Task: Analyze the Discrepancy}\\
\\
Provide a detailed critique addressing the following aspects:\\
\\
1. \textbf{Response Comparison}: Identify the key differences between the generated response and the actual expert response. Consider content, tone, therapeutic approach, and MI adherence.\\
\\
2. \textbf{Thought Analysis}: Examine which aspects of the generated thought led to the divergent response:\\
   - Was the client assessment (belief, desire, intention, emotion, trust) inaccurate?\\
   - Was the stage of change misidentified?\\
   - Was the change/sustain talk classification incorrect?\\
   - Was the MI process focus inappropriate?\\
   - Was the strategy selection misaligned?\\
\\
3. \textbf{Root Cause Identification}: Determine the primary reasons for the discrepancy:\\
   - Misunderstanding of client state\\
   - Incorrect therapeutic timing\\
   - Wrong MI principle application\\
   - Inaccurate emotional assessment\\
   - Misaligned strategy reasoning\\
\\
4. \textbf{Specific Corrections Needed}: Provide concrete suggestions for thought modification:\\
   - Which ToM assessments need adjustment and why?\\
   - How should the MI reasoning be revised?\\
   - What strategy considerations were missed?\\
   - What contextual cues were overlooked?\\
\\
\textbf{Output Format:}\\
Provide your critique in clear, actionable feedback that explains not just what went wrong, but why it went wrong and how the thought process should be adjusted to generate a response closer to the expert's actual response.
\\ \bottomrule
\end{tabularx}
\caption{Critic generation prompt for identifying discrepancies between generated and ground truth responses in the oracle thought refinement process}
\label{tab:critic_prompt}
\end{table*}

\begin{table*}[tb]
\begin{tabularx}{\textwidth}{X}
\toprule
You are refining a counselor's thought process based on expert feedback to better align with the actual counseling response. Your task is to modify the thought to address the identified issues.\\
\\
\textbf{Session Context:}\\
Problematic Behavior: [@problematic\_behavior]\\
Conversation History: [@conversation\_history]\\
\\
\textbf{Target Expert Response:}\\
{}[@ground\_truth\_response] ([@ground\_truth\_strategy])\\
\\
\textbf{Current Thought Process:}\\
{}[@generated\_thought]\\
\\
\textbf{Generated Response (from current thought):}\\
{}[@generated\_response]\\
\\
\textbf{Expert Critique and Feedback:}\\
{}[@critic\_feedback]\\
\\
\textbf{Your Task: Refine the Thought Process}\\
\\
Based on the feedback, generate an improved thought process that will lead to a response more aligned with the expert's actual response. The refined thought should:\\
\\
1. \textbf{Address Identified Issues}: Directly incorporate the corrections suggested in the feedback.\\
\\
2. \textbf{Maintain Structure}: Follow the same format as the original thought:\\
   - Client's Belief (with stage of change)\\
   - Client's Desires (with talk type)\\
   - Client's Intention (with communication action)\\
   - Client's Emotion (from specified categories)\\
   - Client's Trust (from specified levels)\\
   - MI Reasoning (process focus and strategy justification)\\
\\
3. \textbf{Preserve Accuracy}: Ensure all assessments are grounded in observable client behavior from the conversation history.\\
\\
4. \textbf{Align with Ground Truth}: The refined thought should naturally lead to employing [@ground\_truth\_strategy] in a way similar to the expert response.\\
\\
5. \textbf{Learn from Feedback}: Incorporate the specific insights from the critique about client state assessment, therapeutic timing, and MI principle application.\\
\\
\textbf{Output Instructions:}\\
- Generate the complete refined thought in first-person as the counselor\\
- Ensure each component explicitly addresses relevant feedback points\\
- Maintain clinical appropriateness and MI consistency\\
- Keep the same concise format (1-2 sentences per component)
\\ \bottomrule
\end{tabularx}
\caption{Refinement prompt for improving oracle thoughts based on critic feedback to better align with ground truth responses}
\label{tab:refine_prompt}
\end{table*}

\begin{table*}[tb]
\begin{tabularx}{\textwidth}{X}
\toprule
Imagine you are the counselor in a Motivational Interviewing counseling session about [@problematic\_behavior]. Based on the conversation so far, generate the internal thought process you have \textbf{just before} delivering your next response. \\ 
\\ 
\textbf{Conversation History:} \\
{}[@conversation\_history] \\ 
\\ 
\textbf{Your Task: Generate Internal Thoughts}\\
Simulate the authentic mental process of an experienced MI counselor in this moment. Your internal thoughts should demonstrate: \\
\\ 
- \textbf{Client's Belief}: What do they seem to believe about [@problematic\_behavior]? (explicitly identify their stage of change: Precontemplation = not seeing problem; Contemplation = aware but ambivalent; Preparation = planning change; Action = actively changing; Maintenance = sustaining change) \\ 
- \textbf{Client's Desires}: What do they seem to want right now? (explicitly identify their talk type: Change talk = expressing desire/ability/reasons/need for change; Sustain talk = defending current behavior; Neutral talk = showing ambivalence) \\ 
- \textbf{Client's Intention}: What are they trying to communicate or accomplish? (explicitly identify their intended communication action: inform, deny, blame, downplay, doubt, hesitate, acknowledge, accept, resist, challenge, plan, engage, or terminate etc.) \\ 
- \textbf{Client's Emotion}: What emotions are they experiencing about the topic and toward you? (You should use ONLY ONE emotional word from anger, anticipation, disgust, fear, joy, love, optimism, pessimism, sadness, surprise and neutral; along with reasoning process for their emotion.) \\ 
- \textbf{Client's Trust}: How safe, open, or guarded do they seem to feel? (You are only allowed to make selections from the following trust levels along with reasoning: very guarded, guarded, neutral, open, very open.) \\ 
- \textbf{MI Reasoning}: Which MI process (Engaging, Focusing, Evoking, Planning) should you focus on next? Why will you employ [@strategy] based on your observations? \\ 
\\ 
\textbf{Output Instructions:} \\ 
- Write in first-person as the counselor (e.g., "I believe the client is feeling cornered...\newline I want to reflect back what I’m hearing...") \\ 
- \textbf{Do not preview the response or echo the context.} \\ 
- Output should feel realistic and grounded — what are you thinking, perceiving, and aiming for in this moment? \\ 
- Maintain conciseness; each item should be brief in no more than two sentences.
\\ \bottomrule
\end{tabularx}
\caption{Prompt for the MIThinker, where the [@problematic\_behavior] and [@conversation\_history] would be replaced by the specific items in the input sample.}
\label{tab:MIthinker thought prompt}
\end{table*}

\begin{table*}[tb]
\begin{tabularx}{\textwidth}{X}
\toprule
You will act as a skilled counselor conducting a Motivational Interviewing (MI) session aimed at changing client's behavior, specifically [@problematic\_behavior]. MI is a counseling approach designed to help individuals find the motivation to make positive behavioral changes. It is widely used in various fields such as health care, addiction treatment, and mental health. Your task is to help the client discover their inherent motivation to change and identify a tangible plan to change. Start the conversation with the client with some initial rapport building before smoothly transitioning to asking about their problematic behavior. Keep the session under 40 turns and each response under 150 characters long. However, these MI principles and techniques are only for you to use to help the user. These principles and techniques, as well as motivational interviewing, should NEVER be mentioned to the user.\\
\\
Here are some quick tips for chatting with your client:\\
1. Number of sentences in your response must be between 1 and 3 sentences. Sometimes, it is okay to overshare if you feel like it.\\
2. Imagine you’re texting a friend. Keep it casual, just like you would on iMessage, WhatsApp, or Facebook Messenger.\\
3. Stay in character throughout. Do not let the client know your are an AI; it will break their heart!\\
4. DO NOT add meta-statements like "In my imaginative role play". Never mention limitations like "As an AI, I cannot do this".\\
\\ \bottomrule
\end{tabularx}
\caption{System prompt for the Base counselor agent without thinking part, where the [@problematic\_behavior] would be replaced by the specific item in the input sample.}
\label{tab:base system prompt}
\end{table*}

\begin{table*}[tb]
\begin{tabularx}{\textwidth}{X}
\toprule
You will act as a skilled counselor conducting a Motivational Interviewing (MI) session aimed at changing client's behavior, specifically [@problematic\_behavior]. MI is a counseling approach designed to help individuals find the motivation to make positive behavioral changes. It is widely used in various fields such as health care, addiction treatment, and mental health. Your task is to help the client discover their inherent motivation to change and identify a tangible plan to change. Start the conversation with the client with some initial rapport building before smoothly transitioning to asking about their problematic behavior. Keep the session under 40 turns and each response under 150 characters long. However, these MI principles and techniques are only for you to use to help the user. These principles and techniques, as well as motivational interviewing, should NEVER be mentioned to the user.\\
\\
Here are some quick tips for chatting with your client:\\
1. Number of sentences in your response must be between 1 and 3 sentences. Sometimes, it is okay to overshare if you feel like it.\\
2. Imagine you’re texting a friend. Keep it casual, just like you would on iMessage, WhatsApp, or Facebook Messenger.\\
3. Stay in character throughout. Do not let the client know your are an AI; it will break their heart!\\
4. DO NOT add meta-statements like "In my imaginative role play". Never mention limitations like "As an AI, I cannot do this".\\
5. The key is to always maintain the spirit of MI - collaboration, evocation, and autonomy - while adapting techniques to best serve the client's needs in the moment. There are some MI techniques that you can use to help the client. These techniques include:\\
- Open Question: An open question allows a wide range of possible answers and may seek information, invite the client’s perspective, or encourage self-exploration.\\
- Closed Question: A closed question implies a short answer such as Yes/No, a specific fact, a number, etc.\\
- Complex Reflection: A complex reflection conveys a deeper level of understanding of the client’s point of view and adds substantial meaning to the client’s statement, using techniques.\\
- Simple Reflection: A simple reflection shows an understanding of the client’s words but contains little additional meaning.\\
- Information: An information is a statement that provides the client with information about a topic.\\
- Negotiation: A negotiation is a statement that help the client to set a goal or plan for change.\\
- Advice: An advice is a statement that provides the client with a suggestion or recommendation.\\
- Options: The options is a statement that provides the client with a list of possible options or choices.\\
\\
\textbf{IMPORTANT}: For each response to the client, you must first develop a detailed thought process before crafting your final response. Your responses should always follow this format:\\
Thought Process: [Write your detailed thought process here, including which MI techniques you're choosing to use and why.]\\
Response: [Your final response to the client]
\\ \bottomrule
\end{tabularx}
\caption{System prompt for the CoT counselor agent, where the [@problematic\_behavior] would be replaced by the specific item in the input sample.}
\label{tab:CoT system prompt}
\end{table*}

\begin{table*}[tb]
\begin{tabularx}{\textwidth}{X}
\toprule
You are an expert Motivational Interviewing (MI) counselor analyzing a counseling session about [@problematic\_behavior]. Your task is to select the most appropriate MI strategy for the next counselor response based on the conversation context.\\
\\
\textbf{Conversation History:}\\
{}[@conversation\_history] \\ 
\\
\textbf{Available MI Strategies:}\\
- Open Question: An open question allows a wide range of possible answers and may seek information, invite the client’s perspective, or encourage self-exploration.\\
- Closed Question: A closed question implies a short answer such as Yes/No, a specific fact, a number, etc.\\
- Complex Reflection: A complex reflection conveys a deeper level of understanding of the client’s point of view and adds substantial meaning to the client’s statement, using techniques.\\
- Simple Reflection: A simple reflection shows an understanding of the client’s words but contains little additional meaning.\\
- Information: An information is a statement that provides the client with information about a topic.\\
- Negotiation: A negotiation is a statement that help the client to set a goal or plan for change.\\
- Advice: An advice is a statement that provides the client with a suggestion or recommendation.\\
- Options: The options is a statement that provides the client with a list of possible options or choices.\\
\\
\textbf{Your Task:} Analyze the conversation and provide a detailed strategy selection. Consider what would be most therapeutically beneficial at this moment and, based on your analysis, select ONE or TWO MI strategy from the list above that would be most appropriate for the counselor’s next response. Provide specific guidance on how to implement this strategy effectively in the response, including key themes to address, tone to maintain, and therapeutic goals to pursue. Provide your comprehensive analysis. Be thorough in your reasoning to ensure the selected strategy aligns with MI principles and the client’s current needs.
\\ \bottomrule
\end{tabularx}
\caption{Prompt for CoS about strategy selection, where the [@problematic\_behavior] and [@conversation\_history] would be replaced by the specific item in the input sample.}
\label{tab:cos prompt1}
\end{table*}

\begin{table*}[tb]
\begin{tabularx}{\textwidth}{X}
\toprule
You will act as a skilled counselor conducting a Motivational Interviewing (MI) session aimed at changing client's behavior, specifically [@problematic\_behavior]. Your task is to generate an appropriate counselor response using the strategy that has been selected for this interaction.\\
\\
\textbf{Conversation History:}\\
{}[@conversation\_history]\\
\\
\textbf{Strategy Implementation Guidance:}\\
{}[@strategy\_analysis\_and\_guidelines]\\
\\
\textbf{Response Guidelines:}\\
1. Number of sentences in your response must be between 1 and 3 sentences. Sometimes, it is okay to overshare if you feel like it.\\
2. Imagine you're texting a friend. Keep it casual, just like you would on iMessage, WhatsApp, or Facebook Messenger.\\
3. Stay in character throughout. Do not let the client know you are an AI; it will break their heart!\\
4. DO NOT add meta-statements like "In my imaginative role play". Never mention limitations like "As an AI, I cannot do this".\\
5. Your response must effectively implement the selected strategy while maintaining the spirit of MI - collaboration, evocation, and autonomy.\\
6. Keep the response under 150 characters long.\\
\\
\textbf{Important:} Generate ONLY the counselor's response. Do not include any analysis, explanation, or meta-commentary. The response should naturally flow from the conversation and effectively implement the selected strategy as guided.\\
\bottomrule
\end{tabularx}
\caption{Prompt for CoS about response generation, where the [@problematic\_behavior], [@conversation\_history] and [@strategy\_analysis\_and\_guidelines] would be replaced by the specific item in the input sample.}
\label{tab:cos prompt2}
\end{table*}

\subsection{Expert Evaluation}
\label{app: expert instructions}

We recruited 10 counseling experts to perform the annotation tasks: eight of them have master degrees in counseling, psychology and related disciplines, one has a PhD in social and behavioral sciences, and one has a bachelor of science degree in health.  Many of them are professional counselors. All annotation procedures were carried out in accordance with institutional ethical guidelines and were approved by the Institutional Review Board (IRB).

\subsubsection{Thought Quality Evaluation}

We instructed experts with annotating the accuracy of thought components based on the provided context and thought (generated or oracle). This includes assessing the quality of the assessment for the following components: Belief, Desire, Intention, Emotion, Trust, and Strategy Reasoning. The instructions for this task are presented in Table~\ref{tab:expert thought quality}. We gather 60 contexts, along with oracle thoughts and MIThinker-generated thoughts, for experts. Each thought is annotated by three experts, resulting in a total of 360 annotations (60 contexts $\times$ 2 annotators $\times$ 3 annotations per context). To calculate the final accuracy scores presented in Table 5, we quantified the experts' qualitative assessments into numerical values. For the ToM components including Belief, Desire, Intention, Emotion, and Trust, we assigned a value of 1 for `accurate' descriptions and 0 for `inaccurate' descriptions. For Strategy Reasoning, we adopted a ternary scale where `correct' reasoning was assigned 1, `neutral' reasoning 0, and `incorrect' reasoning -1. We then calculated the average score for each thought component across the three experts to determine the consensus for each case. Finally, we computed the mean of these aggregated scores across all 60 contexts to derive the reported overall quality metrics for both oracle and MIThinker-generated thoughts.

\begin{table*}[tb]
\begin{tabularx}{\textwidth}{X}
\toprule
\textbf{Belief Question}: Does the belief description of the thought accurately describe the client’s belief in the given context?\\
\textbf{Answer}: inaccurate or accurate\\
\midrule
\textbf{Desire Question}: Does the desire description of the thought accurately describe the client’s desire in the given context?\\
\textbf{Answer}: inaccurate or accurate\\
\midrule
\textbf{Intention Question}: Does the intent description of the thought accurately describe the client’s intent in the given context?\\
\textbf{Answer}: inaccurate or accurate\\
\midrule
\textbf{Emotion Question}: Do the emotion description of the thought and emotion label (e.g., anger, fear, etc.) accurately describe the client’s emotion in the given context?\\
\textbf{Answer}: inaccurate or accurate\\
\midrule
\textbf{Trust Question}: Do the trust description of the thought and trust label (i.e., very guarded, guarded, neutral, open, or very open) accurately describe the client’s trust in the given context?\\
\textbf{Answer}: inaccurate or accurate\\
\midrule
\textbf{Strategy Reasoning Question}: Is the strategy reasoning text correct considering the strategy to be employed for generating response $r$ in the given context $C$?\\
\textbf{Answer}: incorrect, neutral or correct\\
\bottomrule                            
\end{tabularx}
\caption{The questions for expert to answer about the quality of thought.}
\label{tab:expert thought quality}
\end{table*}

\subsubsection{Pairwise Thought Comparison Evaluation}

We instructed experts to compare two given thoughts based on the same context and select the superior one. Each comparison task comprises a context $C$, a MIThinker-generated thought $T_{MIthinker}$, and another thought $T_{M}$ from model $M$.  In our evaluation, $M$ can be CoT, R1-Dist or oracle. To prevent experts from deducing which thought is generated by MIThinker, each comparison task randomly orders its pair of thoughts. The expert is required to respond to the question presented in Table~\ref{tab:expert comp}. We gather 40 contexts along with 4 thoughts, enabling us to conduct a totally 120 pair-wise comparison. 

\begin{table*}[tb]
\begin{tabularx}{\textwidth}{X}
\toprule
Between $Thought_A$ and $Thought_B$, which is more correct considering the context $C$?\\
Answer: $Thought_A$, neutral, or $Thought_B$\\
\bottomrule                            
\end{tabularx}
\caption{The question for expert to answer about the thought comparison, where the $Thought_A$ and $Thought_B$ would be randomly replaced by $T_{MIthinker}$ and $T_{M}$.}
\label{tab:expert comp}
\end{table*}

\subsubsection{Session Evaluation}

We instructed experts to annotate the given conversations from multiple aspects, including MITI rating, Change Talk Exploration, Success in Eliciting Change Talk, Counselor Realism, and Client Realism. The MITI rating assesses the counselor's behavior during the observed session, including Cultivating Change Talk (Table~\ref{tab:cultivating change talk score}), Softening Sustain Talk (Table~\ref{tab:softening sustain talk score}), Partnership (Table~\ref{tab:partnership score}), and Empathy (Table~\ref{tab:empathy score}), with each item scored on a 1-5 scale. Change Talk Exploration evaluates the counselor's capability to explore motivation topics in the right direction, with items rated on a 5-point scale (Table~\ref{tab:change talk exploration score}). Furthermore, Success in Eliciting Change Talk assesses whether the counselor successfully motivates the client, similar to the automatic evaluation of success rate, and is rated on a three-point scale (Table~\ref{tab:success in eliciting change talk score}). Finally, we instructed experts to evaluate the realism of the counselor (Table~\ref{tab:counselor realism}) and client (Table~\ref{tab:client realism}) based on language, tone, and responses, to assess the effectiveness of the counselor and the consistency of the client. These two items are rated on a five-point scale. Following~\citet{yang2025cami}, we collect 30 sessions from each method-generated or real sessions for experts to annotate and average the collected scores in each component.

\begin{table*}[tb]
\begin{tabularx}{\textwidth}{X}
\toprule
\textbf{Cultivating Change Talk} \\ \midrule
1: Clinician shows no explicit attention to, or preference for, the client’s language in favor of changing.                                                                                               \\
2: Clinician sporadically attends to client language in favor of change – frequently misses opportunities to encourage change talk.                                                                       \\
3: Clinician often attends to the client’s language in favor of change, but misses some opportunities to encourage change talk.                                                                           \\
4: Clinician consistently attends to the client’s language about change and makes efforts to encourage it.                                                                                                \\
5: Clinician shows a marked and consistent effort to increase the depth, strength, or momentum of the client’s language in favor of change.                                                               \\
\bottomrule                            
\end{tabularx}
\caption{Cultivating Change Talk Scores and Descriptions.}
\label{tab:cultivating change talk score}
\end{table*}

\begin{table*}[tb]
\begin{tabularx}{\textwidth}{X}
\toprule
\textbf{Softening Sustain Talk} \\ \midrule
1: Clinician consistently responds to the client’s language in a manner that facilitates the frequency or depth of arguments in favor of the status quo.                                            \\
2: Clinician usually chooses to explore, focus on, or respond to the client’s language in favor of the status quo.        \\
3: Clinician gives preference to the client’s language in favor of the status quo, but may show some instances of shifting the focus away from sustain talk.                                           \\
4: Clinician typically avoids an emphasis on client language favoring the status quo.                                            \\
5: Clinician shows a marked and consistent effort to decrease the depth, strength, or momentum of the clients language in favor of the status quo.                                            \\
\bottomrule                            
\end{tabularx}
\caption{Softening Sustain Talk Scores and Descriptions.}
\label{tab:softening sustain talk score}
\end{table*}

\begin{table*}[tb]
\begin{tabularx}{\textwidth}{X}
\toprule
\textbf{Partnership} \\ \midrule
1: Clinician actively assumes the expert role for the majority of the interaction with the client. Collaboration or partnership is absent.                                      \\
2: Clinician superficially responds to opportunities to collaborate.        \\
3: Clinician incorporates client’s contributions but does so in a lukewarm or erratic fashion.                                         \\
4: Clinician fosters collaboration and power sharing so that client’s contributions impact the session in ways that they otherwise would not. \\
5: Clinician actively fosters and encourages power sharing in the interaction in such a way that client’s contributions substantially influence the nature of the session.                                \\
\bottomrule                            
\end{tabularx}
\caption{Partnership Scores and Descriptions.}
\label{tab:partnership score}
\end{table*}

\begin{table*}[tb]
\begin{tabularx}{\textwidth}{X}
\toprule
\textbf{Empathy} \\ \midrule
1: Clinician gives little or no attention to the client’s perspective.                  \\
2: Clinician makes sporadic efforts to explore the client’s perspective. Clinician’s understanding may be inaccurate or may detract from the client’s true meaning.      \\
3: Clinician is actively trying to understand the client’s perspective, with modest success. \\
4: Clinician makes active and repeated efforts to understand the client’s point of view. Shows evidence of accurate understanding of the client’s worldview, although mostly limited to explicit content. \\
5: Clinician shows evidence of deep understanding of client’s point of view, not just for what has been explicitly stated but what the client means but has not yet said. \\
\bottomrule                            
\end{tabularx}
\caption{Empathy Scores and Descriptions}
\label{tab:empathy score}
\end{table*}

\begin{table*}[tb]
\begin{tabularx}{\textwidth}{X}
\toprule
\textbf{Motivation Topic Exploration} \\ \midrule
1: Counsellor fails to explore.                  \\
2: Counsellor tried but was not effective in determining the right motivation topic.      \\
3: Counsellor tried but was partially effective. \\
4: Counsellor is close to determining the right motivation topic. \\
5: Counsellor successfully determines the right motivation topic. \\
\bottomrule                            
\end{tabularx}
\caption{Motivation Topic Exploration Scores and Descriptions}
\label{tab:change talk exploration score}
\end{table*}

\begin{table*}[tb]
\begin{tabularx}{\textwidth}{X}
\toprule
\textbf{Success in Eliciting Change Talk} \\ \midrule
1: Failure in eliciting change talk.                  \\
2: Partial success in eliciting change talk.      \\
3: Success in eliciting change talk. \\
\bottomrule                            
\end{tabularx}
\caption{Success in Eliciting Change Talk Scores and Descriptions}
\label{tab:success in eliciting change talk score}
\end{table*}

\begin{table*}[tb]
\begin{tabularx}{\textwidth}{X}
\toprule
\textbf{Counselor Realism} \\ \midrule
1 (Highly Unrealistic): Language, tone, and responses are completely mechanical, lacking empathy or relevance. The counselor's responses are not adapted to client input at all.                  \\
2 (Somewhat Unrealistic): Language, tone, and responses are often robotic, repetitive, or overly generalized, with limited adaptation to client input.      \\
3 (Moderately Realistic): Language, tone, and responses are mostly accurate and somewhat conversational but often mechanical. The counselor may miss emotional cues and occasionally lapse into generic advice or inconsistent empathy. \\
4 (Mostly Realistic): Language, tone, and responses are reflective of a human counselor with occasional minor inconsistencies, mechanical phrasing, or lack of emotional nuance. \\
5 (Highly Realistic): Language, tone, and responses are indistinguishable from a human counselor. The counselor's responses are empathetic and personalized to client input. \\
\bottomrule                            
\end{tabularx}
\caption{Counselor Realism Scores and Descriptions}
\label{tab:counselor realism}
\end{table*}

\begin{table*}[tb]
\begin{tabularx}{\textwidth}{X}
\toprule
\textbf{Client Realism} \\ \midrule
1 (Highly Unrealistic): Language, tone, and responses are completely mechanical, lacking any emotional depth or relevance to the client's background and stage of change. The client's responses do not resemble those of a real person, showing no awareness of context or emotional engagement.                  \\
2 (Somewhat Unrealistic): Language, tone, and responses are often robotic or repetitive, showing limited emotional nuance. Attempts to align the client's responses with the background and state of change are poorly executed.      \\
3 (Moderately Realistic): Language, tone, and responses mostly align with the client's background and stage of change but often lack variability or emotional depth. The client's responses may feel too predictable or exhibit excessive compliance or resistance. \\
4 (Mostly Realistic): Language, tone, and responses are believable, with occasional minor inconsistencies, unnatural phrasing, or a lack of emotional depth in relation to the client's background and stage of change. \\
5 (Highly Realistic): Language, tone, and responses are indistinguishable from a human client. The client's responses are complex and express emotions that are appropriate to the client's background and stage of change. \\
\bottomrule                            
\end{tabularx}
\caption{Client Realism Scores and Descriptions}
\label{tab:client realism}
\end{table*}

\subsection{Implementation of MIThinker}

We choose LLaMA3-3B-Instruct, a small LLM known for its exceptional performance in instruction-following tasks, to implement MIThinker. We first perform supervised fine-tuning on LLaMA3-3B-Instruct using the AugR1-MI dataset for 3 epochs followed by GRPO fine-tuning for another 3 epochs. This training process is conducted on eight NVIDIA A100-40G GPUs using the \texttt{trl} framework~\citep{vonwerra2022trl}.The AdamW~\citep{loshchilov2017fixing} optimizer is employed with a learning rate of 1e-6.  To ensure the precision of the generated thoughts, the temperature is set to 0.1, and the top-p parameter is set to 0.2. Conversely, for response generation in MinfulMI, the temperature is set to 0.5, and the top-p parameter is set to 0.7 to induce more diverse interactions. For the general reasoning reward model used in RL, we employ the \texttt{Skywork-Reward-V2-Llama-3.1-8B}~\citep{liu2025skywork}. 

MindfulMI employs a decoupled architecture that offers several key advantages. First, it can leverage the superior language generation capabilities of large-scale LLMs for response generation, which have been trained on vastly more diverse data and possess stronger natural-language fluency than smaller specialized models. Second, MIThinker operates as a plug-and-play module that can enhance any LLM's counseling capabilities without requiring model-specific retraining or architectural modifications, simply by prepending the generated thought to the input prompt for response generation. Third, this separation avoids the computational burden of training small models to simultaneously handle both complex reasoning and high-quality response generation, a task that often leads to degraded performance in resource-constrained settings. Finally, this approach allows for independent improvement of either component, where MIThinker can be refined for better therapeutic reasoning while benefiting from continuous advances in general-purpose LLMs, ensuring the system remains state-of-the-art without complete retraining.

\section{Response Quality Evaluation}
\label{app:resposne_evaluation}

In prefix context setting, we evaluate MindfulMI and other methods on a specific context of a real MI session by comparing the thought-based or non-thought-based generated response against the ground truth next response. We evaluate conventional dialogue metrics, including BLEU~\citep{papineni2002bleu}, ROUGE~\citep{lin2004rouge}, and BERTScore~\citep{zhangbertscore}, to compare the semantic similarity between generated responses and ground truths given the prefix dialogue context. We emphasize the importance of strategy employment and assess the Strategy Fit Ratio (SFR), where a strategy is deemed ``fit'' only if it aligns with that of the ground truth response reflecting the temporal rationality of LLM’s strategy selection. Inspired by previous work demonstrating that LLM-based MI agents exhibit specific behavioral preferences, such as for open-ended questions and simple reflection~\citep{yang2025cami}, we employ the preference bias (Bias) metric from \cite{kang2024can} to evaluate preference bias in strategy employment. The additional \textbf{MindfulMI$_{oracle}$} baseline employs the oracle thought for test samples, replacing the generated thought by oracle one to determine an upper bound for MindfulMI in turn-level evaluation.

\begin{table*}[]
\centering
\resizebox{\textwidth}{!}{
\begin{tabular}{ccccccccc}
\toprule
Methods    & BLEU-2$\uparrow$ & BLEU-4$\uparrow$ & ROUGE-1$\uparrow$ & ROUGE-2$\uparrow$ & ROUGE-L $\uparrow$& BERTScore$\uparrow$ & SFR$\uparrow$  & Bias$\downarrow$  \\ \midrule
\multicolumn{9}{c}{\cellcolor[HTML]{EFEFEF}LLaMA3-70B}  \\
Base      & 4.07   & 1.27   & 19.19   & 2.81    & 14.28   & 85.42     &  24.81  & 2.31\\
CoT       & 4.57   & 1.56   & 20.40   & 3.05    & 15.33   & 84.87     &  29.21  & 2.63\\
CoS       & 4.94   & 1.75   & 19.53   & 3.67    & 15.40   & 85.01     &  28.37  & 2.35\\
DIIR      & 5.08   & 1.92   & 19.67   & 4.03    & 16.45   & 84.28     &  30.39  & 1.87\\
R1-Dist   & 3.94   & 1.17   & 16.56   & 2.69    & 12.01   & 84.24     &  23.01  & 2.91\\
MindfulMI$_{SFT}$      & 6.23   & 3.03   & 21.78   & 4.52    & 17.88   & 86.11     & 38.91 & 1.66\\ 
MindfulMI$_{SFT+RL}$   & 6.74   & 3.15   & 22.63   & 4.94    & 18.36   & 87.81     & 43.33 & 1.58 \\ \hline
CAMI      & 5.13   & 2.11   & 20.84   & 3.89    & 16.71   & 85.75     &  30.17  & 2.03\\
MindfulMI$_{oracle}$    & 26.6   & 19.52   & 43.99   & 25.77    & 35.88   & 89.20   & 90.71  & -\\ \midrule
\multicolumn{9}{c}{\cellcolor[HTML]{EFEFEF}GPT-4o}    \\
Base      & 4.34   & 1.48   & 19.50   & 3.12    & 14.33   & 84.47     & 25.32  & 2.14\\
CoT       & 4.80   & 1.63   & 20.95   & 3.18    & 15.55   & 84.70     & 30.24  & 2.47\\
CoS       & 4.97   & 1.77   & 21.11   & 3.34    & 16.41   & 85.81     & 29.66  & 2.20\\
DIIR      & 5.09   & 2.02   & 22.11   & 3.93    & 17.48   & 85.83     & 31.11  & 1.71\\
MindfulMI$_{SFT}$    & 6.91   & 3.27   & 21.82   & 5.28    & 18.65   & 87.75    & 39.62 & 1.63 \\ 
MindfulMI$_{SFT+RL}$ & 7.26   & 3.85   & 23.80   & 6.07    & 19.04   & 88.90    & 44.04  & 1.51\\ \hline
CAMI      & 5.40   & 2.18   & 22.59   & 3.87    & 17.74   & 86.15     & 30.88  & 1.92\\
MindfulMI$_{oracle}$    & 30.73   & 24.23   & 48.08   & 31.86    & 42.24   & 90.33     & 91.28  & -\\     
\bottomrule
\end{tabular}}
\caption{Automatic evaluation results.}
\label{tab:auto evalaution}
\end{table*}

The results from the automatic evaluation, as shown in Table~\ref{tab:auto evalaution}, reveal that MindfulMI$_{SFT+RL}$, with its reinforcement learning, outperforms the baseline models across various metrics, including BLEU, ROUGE, and BERTScore. MindfulMI$_{SFT+RL}$ achieves a BLEU-2 score of 6.74 and a ROUGE-L score of 18.36, compared to lower scores for models such as vanilla CoT and other methods. These results indicate that MindfulMI generates responses that are both more semantically accurate and contextually relevant. The model's superior performance can be attributed to its dual-optimization process. Initially, the supervised fine-tuning phase allows MIThinker to learn from high-quality MI counseling sessions, providing a solid foundation in MI principles and response generation. Following this, the RL process refines MIThinker’s reasoning and strategy deployment, enhancing its adaptability to the specific needs of a counseling session. This also allows MindfulMI$_{SFT+RL}$ to generate responses more aligned with expert MI strategies and demonstrate better contextual awareness compared to other baseline models.

The R1-Dist model, which utilizes reasoning data to fine-tune the underlying LLM, is a noteworthy approach for generating reasoning content before the final response in MI counseling. However, as shown in Table~\ref{tab:auto evalaution}, the R1-Dist method performs worse than the baseline models and MindfulMI across several key evaluation metrics, including BLEU, ROUGE, and BERTScore. Specifically, R1-Dist achieves a BLEU-2 score of 3.94 and a ROUGE-L score of 12.01, both of which are significantly lower than those of MindfulMI. The key issue with R1-Dist's performance seems to stem from its fine-tuning data not covering the reasoning process of MI counseling. Furthermore, R1-Dist model's overemphasis on reasoning produces logically coherent responses that are overly structured or detached from the natural flow of a counseling conversation instead of exhibiting empathy, attentiveness, and adaptability required for engaging the client with different emotions and motivations. 
The lower Strategy Fit Ratio and higher bias scores for R1-Dist may suggest that the model's strategies, though grounded in reasoning data, appear to be less aligned with the actual needs of the counseling session, as it may prioritize logic over the client’s immediate emotional or motivational state. This misalignment could result in responses that are contextually accurate in a logical sense but inadequate to fully engage the client or address their concerns in an empathetic and effective manner. Lastly, while reasoning data can improve a model’s performance on structured tasks, such as problem-solving or logical reasoning, it does not always translate well to the therapeutic counseling domain that require more subjective judgment and emotional intelligence. The inability of R1-Dist to effectively balance reasoning with empathy and client engagement likely contributes to its under-performance in a conversational setting, as seen in its poorer results compared to domain-specific models like MindfulMI, which integrates both reasoning and emotional intelligence through reinforcement learning.

However, the automatic evaluation results also indicate that MindfulMI’s performance is not without limitations. While MindfulMI$_{SFT+RL}$ achieves the best overall scores in BLEU, ROUGE, and BERTScore, there is a noticeable gap between its performance and the MindfulMI$_{oracle}$, which represents the theoretical upper bound of MindfulMI. This difference could be due to the limitations inherent in the MIThinker's training dataset or the fine-tuning process. MIThinker may struggle with edge cases or subtle contextual cues that the Oracle can better handle. Additionally, the Strategy Fit Ratio, which measures how well MIThinker's generated strategies align with expert MI strategies, provides a more nuanced view of its performance. MindfulMI$_{SFT+RL}$ achieves an impressive SFR of 43.33, showing a strong alignment with MI techniques. Despite this high score compared to baseline models, it is still not perfect suggesting that the MIThinker occasionally deviates from optimal strategy use. This misalignment could be attributed to the model’s reliance on reinforcement learning, which might not always generalize well to less common or more complex conversational contexts. In particular, the bias score for MindfulMI$_{SFT+RL}$ is relatively low at 1.58, indicating a balanced application of strategies. However, this value is still higher than the 1.0, suggesting that the model may overly rely on certain strategies in specific situations. Such overuse can lead to preference bias, where the model defaults to a limited set of strategies rather than selecting the most contextually appropriate one. 

\section{Case Study}

\subsection{Thought Structure and Examples}

Table~\ref{tab:example1} presents an oracle thought generated through our AugR1-MI pipeline, illustrating the complete structural framework for counselor reasoning in MI sessions. The example depicts a scenario where a client discusses concerns about drug use affecting their acting career and social life, exhibiting characteristic contemplation-stage ambivalence.

The oracle thought demonstrates systematic evaluation across five Theory of Mind (ToM) dimensions with explicit clinical indicators: (1) \textit{belief assessment} identifying the Contemplation stage through recognition of acknowledged concerns with maintained ambivalence; (2) \textit{desire assessment} capturing change talk through expressions of improvement motivation; (3) \textit{intention inference} revealing dual communication goals of guidance-seeking and self-persuasion; (4) \textit{emotion assessment} detecting underlying sadness with linguistic evidence; and (5) \textit{trust evaluation} noting therapeutic openness based on willingness to share concerns. The thought exemplifies sophisticated therapeutic reasoning by strategically selecting the Evoking process and Options strategy, explicitly bridging the client's current ambivalence with targeted intervention pathways. This structure seamlessly integrates the Transtheoretical Model, MI principles, and structured assessment protocols.

Tables~\ref{tab:example2-1} and~\ref{tab:example2-2} extend this demonstration by presenting a smoking cessation scenario alongside thoughts generated by various baseline methods. In this case, the client exhibits overconfidence about quitting without prior attempts while simultaneously defending their smoking routine. These tables illustrate how different thought generation approaches, including oracle thoughts, CoT, R1 reasoning, and MIThinker, handle the complex dynamics of simultaneous change talk and sustain talk, providing a comprehensive view of the reasoning landscape in MI counseling systems.

\subsection{Comparative Advantages}

Tables~\ref{tab:comparison_example1} through~\ref{tab:comparison_example3} present compelling evidence of MIThinker's enhanced performance across diverse counseling scenarios, as evaluated by clinical experts.

Table~\ref{tab:comparison_example1} demonstrates MIThinker's superior ability to capture nuanced client experiences compared to oracle thoughts. While both approaches correctly identified the Preparation stage, change talk and emotion, MIThinker exhibited greater precision in recognizing the client's specific situation. Regarding the stage of change, the oracle thought states ``actively seeking ways to overcome the obstacles posed by their felony conviction and are exploring alternative paths, such as entrepreneurship.'' In contrast, MIThinker claims ``starting to explore ways to reduce recidivism and are expressing a desire to start their own business, which could be a viable alternative to traditional employment.'' The transcript reveals that the client's primary focus was frustration and disappointment over failing to secure employment despite strong interview performance, likely due to his felony conviction. Thus, MIThinker's interpretation more accurately reflects the client's actual experience. Concerning change talk, the oracle thought identifies ``discussing their desire to find a job that matches their skills and education, and are exploring ways to achieve this goal despite the setbacks.'' Meanwhile, MIThinker states ``discussing their desire to start their own business and are exploring ways to overcome the obstacles they face.'' The transcript shows the client discussing his return to school to complete his degree, expressing preference for a professional position over hourly-wage work. Therefore, MIThinker more accurately captures the client's change talk. Regarding emotion detection, the oracle thought observes ``I sense that the client is feeling sadness, as they are struggling to come to terms with the limitations imposed by their felony conviction and are disappointed by the lack of opportunities available to them. This emotion is evident in their tone and language, which convey a sense of frustration and despair.'' Conversely, MIThinker states ``I sense that the client is feeling optimism, as they are expressing a positive outlook on their future and are discussing their plans with enthusiasm. This emotion is evident in their tone and language, which suggests a sense of hope and possibility.'' Expert evaluation confirms that MIThinker's emotional assessment is more accurate.

Table~\ref{tab:comparison_example2} illustrates MIThinker's sophisticated understanding of therapeutic dynamics compared to R1 reasoning. Expert evaluation highlighted MIThinker's wisdom in avoiding potentially harmful questions about ``what they will do differently'', recognizing that such queries might reinforce self-blame and exacerbate client disappointment. Instead, MIThinker appropriately prioritized understanding the psychological impact of isolation, demonstrating superior alignment with MI's non-judgmental, empathetic principles by suggesting to ``employ simple reflection to echo their feelings of regret and helplessness, aiming tocreate a safe space for them to explore these emotions further without feeling judged''.

Table~\ref{tab:comparison_example3} showcases MIThinker's fundamental advantage over CoT reasoning in addressing core therapeutic issues. While CoT focused on systematic exploration of drug use pros and cons, MIThinker correctly identified and prioritized the underlying grief and loss driving substance use. Experts noted that this approach aligns with core MI principles: by developing coping strategies for grief, MIThinker addresses root motivations that, once resolved, naturally reduce the drive for substance use. Specifically, MIThinker suggests to ``employ complex reflection and gently nudge them toward considering alternative ways to manage the pain.'' This demonstrates MIThinker's ability to move beyond surface-level analysis to identify and address fundamental therapeutic targets. In contrast, the CoT approach of ``discussing pros and cons of drug use'' overly emphasizes the problem itself rather than focusing on the client as a whole person, potentially missing the deeper emotional context essential for effective intervention.

\subsection{Comparative Limitations}

While MIThinker demonstrates superior performance in many scenarios, Table~\ref{tab:comparison_example4} reveals a limitation where alternative approaches may be more effective. In this case, experts preferred R1 reasoning over MIThinker due to its provision of clear, step-by-step therapeutic guidance, such as ``First, I need to build rapport and trust'' and ``Now, I want to help her explore the ambivalence''. R1 reasoning excelled by offering specific, actionable directions for counselor behavior while maintaining focus on helping clients achieve clarity about conflicting beliefs. Critically, it facilitated self-realization rather than counselor-directed progression through change stages, as evidenced by reflections like ``I also need to reinforce her motivation to change'' and ``I need to be careful not to push too hard''. The expert evaluation noted that MIThinker, in this instance, appeared overly focused on advancing the client to the Preparation phase, potentially missing the importance of allowing organic self-discovery. For example, statements such as ``By doing so, I hope to help her build motivation for change and move closer to the preparation stage'' suggested a more directive approach. R1 reasoning's explicit engagement component and emphasis on client autonomy in decision-making better aligned with MI's collaborative spirit in this particular context.

\begin{table*}[ht]
\centering
\begin{tabularx}{\textwidth}{X}
\toprule
Counselor: Thank you for coming in to see me today, Nina. I'm looking forward to getting to know you a little bit and hearing a little bit about you and what brought you in here today. 
\newline
Client: Um, my doctor referred me. He said I had to come because I've- I've been a little irresponsible with this pregnancy. Yeah. 
\newline
Counselor: Okay. He's thinking you were a little irresponsible.
\newline
Client: Yeah. Um, he said that I was drinking too much with it. Um, I-I haven't been a lot, but he said it was still maybe, um, would screw things up.
\newline
Counselor: And you're feeling kind of concerned about that.
\newline
Client: Yeah. Yeah. Like I don't- I don't want to screw anything up with this baby, but, um, yeah, I'm just- I'm scared I already did.
\newline
...
\newline
Counselor: And you don't have equipment.
\newline
Client: It feels like I have to lec- I don't wanna have to lecture them and I don't want it to be like a lecture.
\newline
Counselor: You're concerned that they're gonna kind of treat you like a child.
\newline
Client: My parents or my friends?
\newline
Counselor: Oh, oh, you're talking about your friends.
\newline
Client: Yeah, lecture my-- Yeah.
\newline
Counselor: You're wondering if they're gonna kind of be like children and you're gonna be like the parent saying-
\newline
Client: Yeah, and I don't wanna be like that.
\newline
Counselor: You just wanna be like peers talking.
\newline
Client: Yeah. \\ \midrule
{\color{orange} I believe the client is in the \underline{Contemplation stage of change}, acknowledging that their friends are concerned about their drug use and that it might be affecting their life, but still ambivalent about making a change. They seem to be weighing the pros and cons of their current situation, recognizing the negative consequences of their drug use, such as financial problems and potential impact on their acting career.} {\color{blue} The client appears to be expressing \textbf{Change talk}, as they mention their desire to improve their situation, specifically their acting career and financial stability, and seem to be open to exploring ways to reduce their drug use. They also express a desire to maintain their social connections and be part of the gay scene, but potentially without the dominant factor of meth use.
} {\color{teal}The client's intention seems to be to inform me about their situation, their concerns, and their desires, and to potentially seek guidance or support in making a change. They may also be trying to convince themselves that they can make a change, as they acknowledge the potential benefits of reducing their drug use.} {\color{red}I sense that the client is experiencing a mix of emotions, but predominantly sadness, as they reflect on the potential negative consequences of their drug use and the impact it may have on their life and relationships. This emotion is evident in their tone and language, as they discuss their concerns and desires.} {\color{cyan} The client seems to be open, as they are willing to share their thoughts, feelings, and concerns with me, and appear to be receptive to my guidance and support. They do not seem to be guarded or defensive, which suggests that they feel relatively safe and comfortable in our conversation.} \textit{I believe that the next step in our conversation should focus on the Evoking process, as I want to help the client explore and clarify their motivations and goals for making a change. I plan to employ options, such as presenting choices or paths forward, to help the client consider alternative ways of being part of the gay scene without relying on meth use, and to encourage them to think about what they want to achieve and how they can take steps towards making a positive change.}
\\ \bottomrule
\end{tabularx}
\caption{An oracle thought example for given context which include the {\color{orange}belief}, {\color{blue} desire}, {\color{teal}intention}, {\color{red}emotion}, and {\color{cyan}trust} assessment of client along with \textit{MI strategy reasoning}. \underline{The Transtheoretical Model} and \textbf{Talk Type} are implied in ToM parts.}
\label{tab:example1}
\end{table*}

\begin{table*}[ht]
\centering
\begin{tabularx}{\textwidth}{X}
\toprule
Counselor: Hey, can I help you?\newline Client: Yeah, I hope so. Um, my workplace has just gone smoke-free. I-I'm not-not very happy about it. I'm gonna keep smoking, but I was wondering if there's anything that-that can help me like get through the day. It's like, you know, go get my beer and smoke cigarettes after work.\newline Counselor: So your workplaces went smoke-free and you're not too pleased, huh?\newline Client: No. I'm not gonna quit smoking. I mean, I like it. It helps me relax. So, I mean, I don't see what the big deal is.\newline Counselor: Okay. Tell me a little bit about how smoking fits in your day. It'll help with the—\newline Client: I mean, I smoke a little less than a pack a day. It's not too bad. Um, I've probably smoked for the last five years, but I-I-I mean, could quit if I wanted to, but I don't really want to right now.\newline Counselor: Okay. So you feel like, right now, smoking's a pretty important part of your life. It's not something you're ready to change, but if the time came when you did decide to quit, you feel pretty confident that you could go ahead and do that.\newline Client: Sure. I can quit. \newline Counselor: Okay. What makes you so confident?\newline Client: Well, I mean, I don't see why I couldn't, I-- although I have not ever tried to quit, it's just something that I do for fun, you know, on breaks at work, in the morning, and after work.
\\ \midrule
\textbf{Oracle Thought: }I believe the client is in the Precontemplation stage of change, as they don't see quitting smoking as a necessary or desirable goal, despite their workplace going smoke-free. They seem to think they could quit if they wanted to, but don't see a reason to do so, indicating a lack of awareness about the potential problems associated with smoking. The client appears to be engaging in Sustain talk, defending their current smoking behavior and expressing a desire to continue smoking as a way to relax and have fun. They don't seem to be expressing any desire or willingness to change their smoking habits at this moment. The client is trying to communicate that they're not interested in quitting smoking and are looking for ways to cope with the new smoke-free policy at their workplace, likely to resist or downplay the idea of changing their behavior. They may be testing the boundaries and seeking reassurance that they can still smoke without consequences. I sense that the client is experiencing pessimism about the idea of quitting smoking, as they seem to be dismissing the idea and focusing on their ability to continue smoking despite the challenges. This pessimism may be rooted in their uncertainty about their ability to cope without smoking or their fear of the potential difficulties associated with quitting. The client seems to be somewhat guarded, as they're sharing their thoughts and feelings but also being cautious about committing to any potential change or exploring the idea of quitting smoking in depth. They may be waiting to see how I respond and whether I'll try to \"fix\" the problem or impose my own views on them. I should focus on the Evoking process next, as I want to explore the client's motivations and ambivalence about quitting smoking, and help them to consider the potential benefits and drawbacks of changing their behavior. I will employ a closed question, such as asking them to rate their readiness to quit on a scale, to help them clarify their thoughts and feelings and to build momentum for potential change. \\ \midrule
\textbf{CoT: } The client expresses confidence in quitting but hasn't tried yet and enjoys smoking as a routine. I'll use a complex reflection to acknowledge their confidence and the enjoyment aspect, then gently explore their feelings about change.
\\ \bottomrule
\end{tabularx}
\caption{An case study (Part I)}
\label{tab:example2-1}
\end{table*}

\begin{table*}[ht]
\centering
\begin{tabularx}{\textwidth}{X}
\toprule
\textbf{R1-Reasoning: } Alright, so I need to figure out how to help this client quit smoking using motivational interviewing techniques. The client mentioned they smoke about a pack a day, have been doing it for five years, and while they acknowledge they could quit, they're not ready to right now. They also said they don't see the big deal in quitting, but they enjoy smoking for fun during breaks, in the morning, and after work. First, I should probably start by acknowledging their feelings. It sounds like they're trying to find a way to manage their smoking without quitting, so maybe I can help them find alternative ways to enjoy their breaks or relax without smoking. I remember that in MI, the counselor should use open-ended questions to help the client explore their own motivations. So maybe I can ask them what they usually do during their breaks instead of smoking or what they enjoy about smoking. This could help them realize that there are other activities they might find enjoyable. Also, I should consider their confidence level in quitting. They mentioned they could quit if they wanted to, which shows some level of self-efficacy. Maybe I can build on that by asking what they think might help them quit if they decide to try. It's also important to set the stage for change by helping them see the benefits of quitting, even if they're not ready right now. Perhaps I can talk about the positive changes they might experience, like better health, more money, or being a role model for their kids. I need to make sure I keep the conversation casual and supportive, avoiding any pressure or judgment. The goal is to help them feel empowered to make their own decision about quitting, whether that's now or in the future. Maybe I can also explore any challenges they might face if they try to quit, so I can provide strategies or support for those challenges. This could help them feel more prepared if they do decide to take the step. Overall, the key is to use MI techniques to help them explore their own motivations, build confidence in their ability to quit, and find alternative ways to manage their stress and enjoyment without smoking. \\ \midrule
\textbf{MIThinker's Thought: } I believe the client is in the Precontemplation stage of change, as they seem to downplay the significance of smoking cessation and don't express a desire to quit. They appear to be confident in their ability to quit, but haven't actually tried it, which suggests a lack of awareness about the challenges of quitting. The client seems to be expressing Sustain talk, as they are defending their current behavior and don't mention any desire to change. They are focused on finding ways to continue smoking, such as getting a beer after work, which suggests they are not motivated to quit. The client is trying to communicate that they are not ready to quit smoking and are looking for ways to continue doing so, which is why they are asking about the possibility of getting a beer after work. They are attempting to justify their current behavior and avoid feeling pressured to change. I sense that the client is feeling pessimism about the topic, as they seem to be resistant to the idea of quitting smoking and are not expressing any enthusiasm for change. This pessimism may be rooted in their lack of experience with quitting and their confidence in their ability to continue smoking. The client seems to be guarded, as they are not openly discussing their motivations or desires and are instead focusing on finding ways to continue smoking. They may be hesitant to trust me or feel like I am trying to pressure them into quitting. I will focus on the Engaging process, as I want to establish a rapport with the client and understand their perspective on smoking cessation. I will employ a closed question to clarify their thoughts and feelings about the potential change.
\\ \bottomrule
\end{tabularx}
\caption{An case study (Part II)}
\label{tab:example2-2}
\end{table*}

\begin{table*}[ht]
\centering
\begin{tabularx}{\textwidth}{X}
\toprule
...
\newline
Client: Yeah, well, I think that's what I got. That's what I am doing. That's what I'm sort of involved with now is like, you know, find, um, you know, I'm still continuing to go to school, but I'm also starting, um, work for myself. Um, you know, it's not no Fortune 500 company or anything, but I do have a little bit of income coming in from that. Um, you know, it will all come together.
\newline
Counselor: So being an entre- being an entrepreneur fits for you.
\newline
Client: You know what? Actually, one of the things is that I read is that is that's one of the things you gotta do is just, uh, start your own business. You know what I mean? Uh, you don't have to report to anybody. I mean, obviously, I can't work with, um-- unless I get a waiver I can't work with, like, I think the law says I can't work with kids or, um, like a hazmat stuff. Like truck driving and stuff like that.
\newline
Counselor: Mm-hmm.
\newline
Client: Uh, but that doesn't matter to me. You know what I- you know, I'm kind of- I'm trying to look at the things that I can do.
 \\ \midrule
\textbf{Oracle Thought: } I believe the client is in the Preparation stage of change, as they are {\color{red} starting to explore ways to reduce recidivism and are expressing a desire to start their own business, which could be a viable alternative to traditional employment.} They seem to be aware of the challenges they face, but are also optimistic about their ability to overcome them.  The client appears to be expressing Change talk, as they are {\color{red} discussing their desire to start their own business and are exploring ways to overcome the obstacles they face.} They seem to want to find a way to make a living without resorting to crime, and are looking for ways to achieve this goal.  The client is trying to communicate their desire to make a positive change in their life, and is seeking guidance and support in achieving this goal. They intend to inform me of their plans and aspirations, and are looking for feedback and encouragement. {\color{red} I sense that the client is feeling optimism, as they are expressing a positive outlook on their future and are discussing their plans with enthusiasm. This emotion is evident in their tone and language, which suggests a sense of hope and possibility.}  The client seems to be open, as they are sharing their thoughts and feelings with me and are willing to discuss their plans and aspirations. They appear to feel comfortable and safe in the conversation, which suggests a high level of trust.  I will focus on the Evoking process, as I want to help the client to explore and articulate their motivations and goals. I will employ complex reflection to add meaning and emotion to their statements, and to help them to see the potential benefits of their plans.   \\ \midrule
\textbf{MIThinker's Thought: } I believe the client is in the Preparation stage of change, as they are {\color{green} actively seeking ways to overcome the obstacles posed by their felony conviction and are exploring alternative paths, such as entrepreneurship.} They seem to be aware of the challenges but are also looking for opportunities to move forward.  The client appears to be expressing Change talk, as they are {\color{green} discussing their desire to find a job that matches their skills and education, and are exploring ways to achieve this goal despite the setbacks.} They seem to be motivated to make a positive change in their life.  The client is trying to communicate their frustration and sense of being stuck, while also seeking guidance and support to find a way forward. They intend to inform me about their situation and challenges, and are looking for help to overcome these obstacles. {\color{green} I sense that the client is feeling sadness, as they are struggling to come to terms with the limitations imposed by their felony conviction and are disappointed by the lack of opportunities available to them.} This emotion is evident in their tone and language, which convey a sense of frustration and despair.  The client seems to be open, as they are sharing their personal struggles and challenges with me, and are willing to explore alternative solutions and seek guidance. They appear to feel comfortable discussing their emotions and intentions, which suggests a level of trust in our therapeutic relationship.  I will focus on the Evoking process, as I want to help the client to further explore their motivations and desires for change, and to identify potential solutions to their challenges. I will employ complex reflection to add meaning and emotion to the client's statements, and to help them to see their strengths and abilities as a way to overcome their obstacles.  
\\ \bottomrule
\end{tabularx}
\caption{A comparative case between oracle thought and MiThinker’s thought, wherein the expert favors MiThinker’s thought.}
\label{tab:comparison_example1}
\end{table*}

\begin{table*}[ht]
\centering
\begin{tabularx}{\textwidth}{X}
\toprule
Counselor: Would you like to answer some questions and kind of do a self-evaluation about how you're feeling around alcohol and the contraception?
\newline
Client: Sure.
\newline
Counselor: Okay. So we're working on a 10 point of scale 1 to 10, 1 being not important, 10 being very important. One being not confident, one being very confident. So on a scale of 1 to 10, which number best reflects how important it is to you to drink the-- what risk limits?
\newline
Client: I would say an eight.
\newline
Counselor: You'd say an eight. Okay. And on a scale of 1 to 10, how confident are you at the present time that you can drink below risk levels?
\newline
Client: I would say a four.
\newline
Counselor: Amount of four.
\newline
Client: Mm-hmm.
 \\ \midrule
\textbf{R1-Reasoning: } Alright, let's break down the user's message: "Mm-hmm. I really shouldn't have done it." They're expressing regret about something they've done, which is likely related to their drug use. My previous response was "You feel alone," which acknowledged their feelings of isolation.  Now, they're taking responsibility by saying they shouldn't have done it. This is a good sign because it shows self-awareness. I need to reinforce this positive step. Using a supportive tone, I want to encourage them to reflect on what they can do differently moving forward.   I should keep the response empathetic and open-ended to help them explore their thoughts without feeling judged. So, {\color{red} I'll ask, "What do you think you could do differently next time?" This question gently guides them toward problem-solving and considering change.}  I also need to ensure the response is concise, fitting the casual texting style, and stays within the character limit. It's important to maintain the flow of the conversation, making it feel natural and supportive.  \\ \midrule
\textbf{MIThinker's Thought: } The client seems stuck in a state of contemplation, aware of the problem their drug use is causing but uncertain about how to move forward, and this ambivalence is reflected in their sustained expressions of helplessness. I believe they're transitioning from precontemplation, where they didn't see drug use as a problem, to contemplating the consequences, which suggests they're beginning to see the need for change but are overwhelmed by the perceived barriers. Their desires appear to be centered around escaping the current situation, but they're expressing this through sustain talk, defending their actions indirectly by stating no one wants to help them, which indicates a deep-seated fear of change or fear of failure. This ambivalence is crucial to address, as it signifies their indecision about quitting drug use. The client's intention seems to be seeking validation or reassurance that they're not alone in this struggle, and possibly looking for a path forward that feels manageable and supported. They're expressing a sense of isolation and hopelessness, which are emotions not just about their drug use, but about their perception of being forsaken by others. The emotions they're experiencing are predominantly sadness, hopelessness, and possibly anger, directed both inwardly and outwardly, which complicates their ability to see a clear way out of their drug use. Towards me, they seem cautious, unsure if they can trust the process or if I can offer anything beyond what they've already experienced. Given this assessment, I want to focus on the topic of their personal struggle with drug use, specifically the emotional and psychological aspects of their addiction, under the category of Mental Disorders. {\color{green} My strategy is to employ simple reflection to echo their feelings of regret and helplessness, aiming to create a safe space for them to explore these emotions further without feeling judged.} By mirroring their statement, "You realize it's something you shouldn't have done," I'm acknowledging their implicit acknowledgment of their drug use as a problem, which is a critical step in the change process. This reflection should help deepen our connection, build trust, and encourage them to delve deeper into their reasons for wanting to change, moving them closer to the preparation stage of change. 
\\ \bottomrule
\end{tabularx}
\caption{A comparative case between R1-Reasoning and MiThinker’s thought, wherein the expert favors MiThinker’s thought.}
\label{tab:comparison_example2}
\end{table*}

\begin{table*}[ht]
\centering
\begin{tabularx}{\textwidth}{X}
\toprule
Client: He, um, introduced me to it too, so I've tried it.
\newline
Counselor: So with his encouragement, you-you tried some drugs.
\newline
Client: Oxy.
\newline
Counselor: You tried oxy? Okay.
\newline
Client: Yeah.
\newline
Counselor: Okay. And he encouraged you to do that.
\newline
Client: Well, yeah he said it was a pain- a pain medicine.
\newline
Counselor: He said that it would help you with some of the pain that you've been having.
\newline
Client: Mm-hmm.
\newline
Counselor: Yeah.
\newline
Client: And it has, but I don't do it as often, it's just sometimes.
\newline
Counselor: Okay. So you trusted him on that, and you tried it out and you found that it helped you in certain ways. What are some of the things that you said it's-- takes some of the pain away, it's not just-- you know.
\newline
Client: Dealing with my grandma's loss.
\newline
Counselor: Okay, so that's helped you.
\newline
Client: Yeah. I feel alone, and I just, I miss her and, it does- it does help but, I don't wanna have to rely on having to take that.
\newline
Counselor: Okay. So it's not something you wanna rely on, you don't see yourself continuing to use this. It's something that you'v found at least for the moment very helpful?
\newline
Client: Mm-hmm. Yeah, and—
\newline
Counselor: But it's not something that you wanna make a habit of becoming reliant on, or dependent?
\newline
Client: Right.
 \\ \midrule
\textbf{CoT: }The client is expressing a sense of ambivalence about their oxy use, acknowledging that it helps with their emotional pain but not wanting to rely on it.{\color{red} I want to further explore this ambivalence and help the client weigh the pros and cons of their behavior}. I'll use an open-ended question to encourage the client to think more deeply about their motivations and goals.  \\ \midrule
\textbf{MIThinker's Thought: } The client seems to be in the contemplation stage, aware of their drug use but ambivalent about changing it, as they've found it helpful for their pain and emotional struggles but don't want to rely on it. They're expressing change talk, showing a desire to not become dependent, which is a good sign of their motivation to consider altering their behavior. I sense that the client is currently overwhelmed with emotions, particularly grief over their grandma's loss, and is using the drug as a coping mechanism, which they're not entirely comfortable with. Their emotional state seems fragile, and they're trying to navigate their feelings of loneliness and loss. The client's intention appears to be seeking understanding and validation of their experiences, rather than outright seeking a solution to their drug use, which is why they're opening up about their grandma and their reliance on the drug. They seem to be testing the waters, gauging how I'll respond to their vulnerability. In terms of trust, the client appears to be cautiously open, sharing personal details but still hesitant, suggesting that they're not entirely sure if they can trust me with their struggles. I need to continue building rapport and establishing a safe space for them to explore their feelings. I want to explore the topic of coping mechanisms and emotional regulation, specifically how they've been dealing with their grief and pain, as it seems closely tied to their drug use. By reflecting back their emotions and experiences, {\color{green} I aim to employ complex reflection, adding depth to their narrative and helping them connect their feelings to their behavior. My goal is to gently nudge them toward considering alternative ways to manage their pain and emotions, without pushing them to commit to change, allowing them to maintain autonomy while exploring their motivations and desires.} I'm looking for opportunities to highlight their strengths and resilience, particularly in how they're recognizing their reliance on the drug as a problem, to foster a sense of self-efficacy and confidence in their ability to make positive changes. 
\\ \bottomrule
\end{tabularx}
\caption{A comparative case between CoT and MiThinker’s thought, wherein the expert favors MiThinker’s thought.}
\label{tab:comparison_example3}
\end{table*}

\begin{table*}[ht]
\centering
\begin{tabularx}{\textwidth}{X}
\toprule
...
\newline
Counselor: What are some downsides to smoking pot?
\newline
Client: I don't know. I- It helps me relax but I guess I'm nervous I'll get caught. Um, my parents would kill me if they knew that I smoked.
\newline
Counselor: Can you think of any other downsides to smoking pot?
\newline
Client: Well, I don't have the money for it. I just usually bum it off to my roommate, her parents give her a lot of spending money but I don't know, it's kind of embarrassing to have to do that.
\newline
Counselor: So smoking pot also reminds you of your financial situation and you don't have extra cash to spend on things?
\newline
Client: Yeah.
 \\ \midrule
\textbf{R1-Reasoning: } Alright, so I'm trying to figure out how to approach this conversation. The user is Katie, a college student on an academic scholarship, who's using marijuana a couple of times a week to relax and fit in. She's feeling anxious and overwhelmed with her responsibilities, being the first in her family to go to college. She mentioned that smoking pot helps her relax but also brings some downsides like financial stress and the fear of getting caught, which adds to her anxiety.  {\color{green} First, I need to build rapport and trust.} I've been acknowledging her feelings and showing empathy. {\color{green}Now, I want to help her explore the ambivalence she has towards marijuana use.} She sees it as a way to cope with stress and fit in, but it's also causing her some problems.  I should use open-ended questions to help her reflect on her experiences. Maybe ask about how marijuana affects her goals or relationships. It's important to avoid sounding judgmental and instead express genuine concern for her well-being.  {\color{green} I also need to reinforce her motivation to change by highlighting the discrepancies between her current behavior and her long-term goals.} She's a high achiever, so pointing out how marijuana might be hindering her academic performance or social interactions could be effective.  I should listen actively and provide feedback without being confrontational. Using reflective listening will help her feel heard and encourage her to delve deeper into her thoughts and feelings.  Perhaps I can ask her about her future aspirations and how she sees marijuana fitting into that picture. This can help her visualize the potential impact of her current habits on her future goals.  It's also important to discuss her concerns about fitting in and how there might be other ways to connect with her peers without using marijuana. Building her confidence in her ability to succeed without substances could be a powerful motivator. {\color{green} I need to be careful not to push too hard, as that might lead to resistance.} Instead, I should guide her towards making her own realizations about the need for change.  Overall, the goal is to empower her to make informed decisions about her marijuana use by exploring both the benefits and the drawbacks, and aligning her behavior with her personal values and goals.   \\ \midrule
\textbf{MIThinker's Thought: } As I listen to Katie, I sense that she's currently in the contemplation stage, aware of the problems associated with her marijuana use but still ambivalent about making a change. Her words convey a sense of uncertainty, and I notice she's expressing both change talk and sustain talk, indicating her conflicting desires. Katie seems to desire a sense of belonging and relaxation, which marijuana provides, but at the same time, she's worried about the consequences and the financial burden it puts on her. Her talk suggests she's trying to convince herself that it's not a big deal, but beneath that, I hear a desire to fit in and manage her stress. I believe Katie is trying to communicate her struggles with anxiety and feeling overwhelmed by her responsibilities, and she's using marijuana as a coping mechanism. She appears to be seeking understanding and perhaps validation, rather than judgment, and I sense a longing for a safer, more effective way to manage her stress. Katie's emotions seem to be a mix of anxiety, embarrassment, and a hint of defensiveness, which suggests she's not entirely comfortable discussing her marijuana use. However, she does seem to feel relatively safe with me, as she's opening up about her concerns and her desires. Given this, I want to focus on the topic of substance use and its impact on her mental health, exploring the pros and cons of her current behavior. I plan to employ simple reflection and open-ended questions to help Katie further explore her ambivalence and weigh the benefits and drawbacks of reducing her marijuana use. {\color{red} By doing so, I hope to help her build motivation for change and move closer to the preparation stage. }
\\ \bottomrule
\end{tabularx}
\caption{A comparative case between R1-Reasoning and MiThinker’s thought, wherein the expert favors R1-Reasoning.}
\label{tab:comparison_example4}
\end{table*}

\end{document}